\documentclass[runningheads]{llncs}
\usepackage[T1]{fontenc}
\usepackage{graphicx}
\usepackage{booktabs}
\usepackage[misc]{ifsym}
\newcommand{\corr}{(\Letter)}

\usepackage{amsmath, amssymb, dsfont}
\usepackage{xargs}

\newcommand{\1}[1]{\mathds{1}\lbrace #1 \rbrace\xspace}
\newcommand{\bbE}{\mathop{\mathbb{E}}\xspace}

\renewcommand {\cal}[1]{\ensuremath{{\mathcal{#1}}}\xspace}

\newcommand\norm[1]{\left\lVert#1\right\rVert\xspace}

\newcommand\zeroOne{0\mbox -1\xspace}
\newcommand\loss{\cal L\xspace}
\newcommand{\zeroOneloss}{\loss^{\zeroOne}\xspace}
\newcommand\mixset[1][h]{\mathbf{#1}}
\newcommand\mixweights[1][q]{\mathbf{#1}}
\newcommandx{\mixture}[2][1=\mixset,2=\mixweights]{\mathbf{#1_{#2}}}

\usepackage{url}
\usepackage[ruled,vlined]{algorithm2e}
\usepackage[noend,compatible]{algpseudocode}
\usepackage{subcaption}
\usepackage{xcolor}
\definecolor{ctcolormain}{RGB}{61, 149, 242}
\definecolor{ctcoloraccessory}{RGB}{245, 156, 47}
\usepackage{array}
\usepackage{multirow}
\usepackage{thmtools, thm-restate}
\usepackage{tikz}
\usetikzlibrary {arrows.meta}
\usetikzlibrary {patterns,patterns.meta}
\usetikzlibrary {topaths,calc}
\usetikzlibrary {graphs}
\usetikzlibrary {graphs.standard}

\graphicspath{{images/}{../images/}}

\begin{document}

\title{Lattice Climber Attack: Adversarial attacks for randomized mixtures of classifiers}
\toctitle{Lattice Climber Attack: Adversarial attacks for randomized mixtures of classifiers}

\titlerunning{Adversarial attacks for randomized mixtures of classifiers}

 \author{Lucas Gnecco Heredia \corr \and Benjamin Negrevergne \and Yann Chevaleyre}
\tocauthor{Lucas~Gnecco-Heredia,Benjamin~Negrevergne,Yann~Chevaleyre}


\institute{LAMSADE, CNRS, Université Paris Dauphine - PSL \\ Correspondence to \email{lucas.gnecco-heredia@dauphine.psl.eu}}

\maketitle              

\begin{abstract}
    Finite mixtures of classifiers (a.k.a. randomized ensembles) have been proposed as a way to improve robustness against adversarial attacks. However, existing attacks have been shown to not suit this kind of classifier. In this paper, we discuss the problem of attacking a mixture in a principled way and introduce two desirable properties of attacks based on a geometrical analysis of the problem (effectiveness and maximality). We then show that existing attacks do not meet both of these properties. Finally, we introduce a new attack called {\em lattice climber attack} with theoretical guarantees in the binary linear setting, and demonstrate its performance by conducting experiments on synthetic and real datasets.
    
\end{abstract}

\section{Introduction}

Deep neural networks have been shown to be vulnerable to adversarial attacks \cite{goodfellow2014explaining}, i.e. small perturbations that, although imperceptible to humans, manage to drastically change the predictions of the model. This observation has led to numerous efforts to understand this phenomenon~\cite{bubeck2019adversarial} and started a series of publications introducing various techniques to train robust models~\cite{madry2017towards} as well as new algorithms to attack them~\cite{tramer2020adaptive}.

One research direction that has been explored is the use of \textit{randomized classifiers}, which include a source of randomness in their predictions. Examples of such classifiers include stochastic pruning~\cite{dhillon2018stochastic,panousis2021stochastic}, noise injection classifiers~\cite{he2019parametric}, classifiers with random input transformations~\cite{xie2017mitigating,raff2019barrage}, finite mixtures~\cite{pinot2020randomization,meunier2021mixed}, among others. Unfortunately, the robustness of these randomized classifiers is not well understood, and most of them have been shown to be less robust than originally claimed under the white-box threat model. The classifier based on random input transformations by Xie et al.~\cite{xie2017mitigating}, which won the 2017 Neurips \emph{Adversarial Attacks and Defences} Competition, was broken by Athalye et al.~\cite{athalye2018obfuscated}, together with stochastic pruning \cite{dhillon2018stochastic}. The defense by Panousis et al.~\cite{panousis2021stochastic} was debated on a Github issue\footnote{Available at \url{https://github.com/fra31/auto-attack/issues/58}} by the authors of Robustbench \cite{croce2020robustbench}, who found that the classifier was not robust using a simple adaptation of AutoAttack \cite{croce2020robustbench}. The defense \emph{Barrage of random transforms} by Raff et al.~\cite{raff2019barrage}, which had impressive robustness results on Imagenet, was broken three years later by Sitawarin et al.~\cite{sitawarin2022demystifying}, and Dbouk et al.~\cite{dbouk2022adversarial} debate the robustness of finite mixtures.

In addition to showing that the real robustness of randomized classifiers is not yet fully understood, these results also highlight the lack of adaptive attacks for randomized models. The attacks used to evaluate the robustness of these models are often not suitable for the task, leading to an overestimation of their robustness, a phenomenon that is well known nowadays~\cite{athalye2018obfuscated,tramer2020adaptive}. 
The lack of strong adaptive attacks for randomized classifiers has undermined research on this family of classifiers and limited its practical applications.
As an example, one of the criteria used by the Robustbench benchmark to filter out defenses is the use of randomness in the forward pass, because such defenses often ``only make gradient-based attacks harder but do not substantially improve robustness''~\cite{croce2020robustbench}.

Arguably, finite mixtures of classifiers \cite{pinot2020randomization,meunier2021mixed} are one of the simplest kind of randomized classifiers, and yet they are not trivial to attack. Dbouk et al.~\cite{dbouk2022adversarial} showed that the adaptations of projected gradient descent (PGD)~\cite{madry2017towards} used by Pinot et al.~\cite{pinot2020randomization} or Meunier et al.~\cite{meunier2021mixed} were weak, thus overestimating the robustness of finite mixtures. They design ARC, the current state-of-the-art attack for finite mixtures of classifiers, and their evaluation shows a considerable drop in robustness with respect to the results reported by Pinot et al.~\cite{pinot2020randomization}. 

In this work, we take a principled approach to understand adversarial attacks for finite mixtures of classifiers using a set-theoretic perspective and the concept of {\em vulnerability regions}. 
We show that the problem of attacking a finite mixture can be seen as the problem of climbing a lattice. Using this perspective, we identify a series of desirable properties and limitations of existing attacks.
Afterward, we leverage the lattice reformulation to devise a new attack with better theoretical guarantees in binary classification with linear classifiers. More specifically, we make the following 3 contributions: First in Section~\ref{sec:attack:failures-existing-attacks} we model the problem of attacking a mixture using a semi-lattice, which allows us to better characterize the limitations of existing attacks like adaptations of PGD~\cite{madry2017towards} and ARC~\cite{dbouk2022adversarial}. Second, we propose in Section~\ref{sec:attack:lca} a new attack algorithm called \textit{lattice climber} that has strong guarantees in the binary linear setting compared to existing attacks. We then generalize to multiclass differentiable classifiers like neural networks. Finally, we provide extensive experimental results showing that our proposed attack is better at simultaneously attacking a finite mixture compared to existing attacks. Our code is available in \url{https://github.com/lucasgneccoh/lattice_climber_attack}.

\section{Preliminaries}

\subsubsection{Notations. } 
For a predicate $C$, we denote by $\1{C}$ the function that returns 1 if the predicate $C$ is true and 0 otherwise. For an integer $m$, we use the notation $\left[m\right]=\{1,\cdots ,m\}$. For a vector $u \in \mathbb{R}^d$, we denote by $u^{(j)}$ the $j$-th component of $u$.
We denote by $\Delta^n$ the probability simplex in $\mathbb{R}^n$.
For a probability vector $p~\in~\Delta^n$, we denote $\mathrm{Cat}(p)$ the categorical distribution on $n$ elements, where $p^{(i)}$ is the probability of sampling element $i$. To alleviate the notation, when $z$ is a random variable following the distribution $\mathrm{Cat}(p)$, we write $z \sim p$.

\subsubsection{Problem setting.}
Given a $d\mbox -$dimensional input space $\cal X \subset \mathbb R^d$ and a set $\cal Y = [K]$ of $K$ class labels, a \textit{deterministic} classifier $h : \cal X \to \cal Y$ is a function that maps each input point $x$ to a predicted label $h(x)$. To measure the quality of the prediction of $h$ at a point $(x,y) \in \cal X \times \cal Y$, we use the zero-one loss :
\begin{equation}\label{eq:zero_one_loss}
    \zeroOneloss(h, x, y) = \1{h(x) \neq y}
\end{equation}

In this paper, we consider finite mixtures of classifiers \cite{pinot2020randomization,meunier2021mixed}, which are a type of randomized classifier inspired by mixed strategies in game theory. Given a base set of deterministic classifiers $\mixset = \{h_1, \dots ,h_m\}$ and a probability distribution over them $\mixweights \in \Delta^m$,  the mixture classifier $\mixture$ will map any input $x$ by first sampling a classifier $i \sim \mixweights$, and then returning $h_i(x)$. Note that, unlike classical deterministic classifiers, mixtures may predict different labels for the same input $x$ over repeated calls, thus the prediction $\mixture(x)$ is a random variable $\hat{y}$ over $\cal Y$, and the zero-one loss needs to be adapted to measure the expected error:
\begin{equation}\label{eq:zero_one_mixture}
    \zeroOneloss(\mixture, x, y) = \bbE_{i \sim \mixweights} [\zeroOneloss(h_i, x, y)] = \sum_{i=1}^{m} \mixweights^{(i)} \cdot \zeroOneloss(h_i, x, y).
\end{equation}
In other words, the zero-one loss of a mixture represents the {\em probability} of predicting an incorrect label for input $x$.


\subsubsection{Adversarial attacks on classifiers and mixtures.}
Given an input point $x \in \cal X$ and its true label $y$, attacking a classifier $h$ (deterministic or randomized) consists of crafting a norm bounded perturbation $\delta \in \mathbb R^d$ (with $\norm{\delta} \le \epsilon$) that increases the $\zeroOne$ loss at $(x, y)$. Various norms can be used to measure the magnitude of the perturbation $\delta$, the most common being $\ell_p$ norms with $p = 2$ or $p = \infty$. For a given $p$-norm, we denote $B_p(x, \epsilon)$ the ball centered at $x$ with radius $\epsilon$ \textit{i.e.} $B_p(x, \epsilon) = \{ x + \delta \in \cal X ~s.t.~ \norm{\delta} \le \epsilon \}$.
The adversarial zero-one loss $\zeroOneloss_{\epsilon}$ is defined as the zero-one loss under attack by an \textit{optimal adversary}: 
\begin{equation}
    \zeroOneloss_{\epsilon,p}(h, x, y) = \sup\limits_{x' 
    \in B_p(x, \epsilon)} \zeroOneloss(h, x', y)
\end{equation}

\section{Failures of existing attacks} \label{sec:attack:failures-existing-attacks}


We start by analyzing of the limitations of existing attacks such as Expectation Over Transformation (EOT) \cite{athalye2018synthesizing} and  ARC \cite{dbouk2022adversarial}. 

\subsection{Attacks based on Expectation Over Transformation (EOT)}

\emph{Expectation Over Transformation} (EOT) was initially developed by Athalye et al.~\cite{athalye2018synthesizing}, in order to craft attacks that are robust to real world perturbations. EOT introduces a set of  transformations $T$ that may be applied to the input, and optimizes the expected loss $\loss$  over $T$:
\begin{equation} \label{eq:attack:eot}
  \displaystyle \bbE_{t \sim T}\left[\loss(h, t(x'), y)\right].
\end{equation}
The EOT principle was later adapted to attack finite mixture of classifiers by Pinot et al.~\cite{pinot2020randomization}. Instead of considering random transformations, the idea is to account for the sampling of the classifier as the source of randomness. To attack the mixture $\mixture$, our objective function becomes the \emph{Expectation Of the Loss} (EOL), as follows:
\begin{equation} \label{eq:attack:eot-finite-mixture}
  \displaystyle \bbE_{i \sim \mixweights}\left[\loss(h_i, x, y)\right] =  ~\displaystyle \sum_{i = 1}^{m} ~ \mixweights^{(i)} ~ \loss(h_i, x, y),
\end{equation}
Note that if we choose $\loss$ to be the zero-one loss, then maximizing the objective in \eqref{eq:attack:eot-finite-mixture} directly corresponds to maximizing the classification error of the mixture in \eqref{eq:zero_one_mixture}. Therefore, \emph{maximizing the EOL with the zero-one loss is the correct objective for attacking a finite mixture}. However, as we will demonstrate, this stops being true if we replace the zero-one loss with a surrogate loss function such as the cross-entropy loss.

\begin{figure}[!ht]
    \begin{minipage}{0.4\textwidth}
        \centering
        \includegraphics[width=0.9\linewidth]{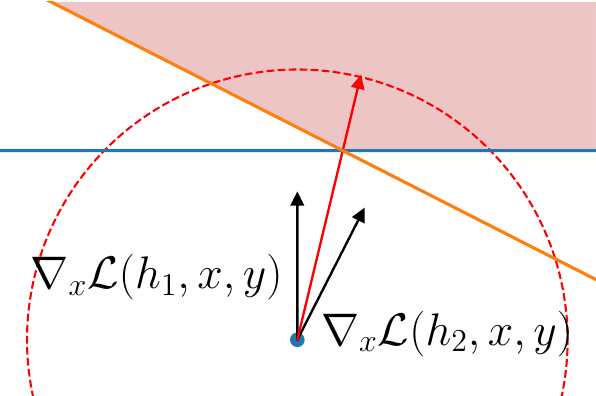}
    \end{minipage}\hfill
    \begin{minipage}{0.4\textwidth}
        \centering
        \includegraphics[width=0.9\linewidth]{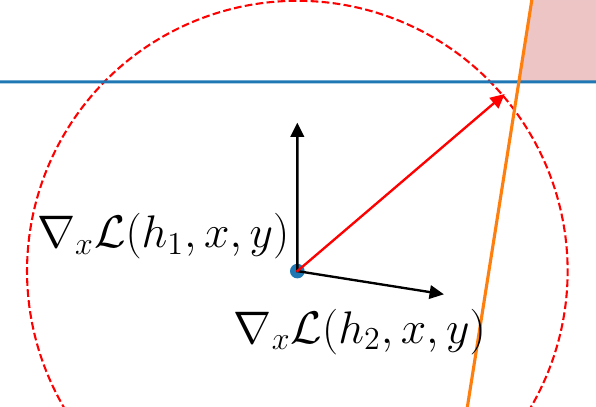}
    \end{minipage}\hfill
    \caption{Example of two linear classifiers with high (left) and low (right) alignment. The red arrow represents the gradient of the loss of the mixture, while the black arrows represent the gradient of the loss of individual classifiers. The red region in the top-right corner represents the points that are adversarial for both classifiers simultaneously. In the case of high alignment (left), the linear combination of the individual gradients is a successful attack direction, while in the case of low alignment (right), it is \emph{not} an effective attack direction as it leads to a non-adversarial perturbation, even though adversarial perturbations do exist.}
  \label{fig:attack:alignment-high-low}
\end{figure}

\paragraph{Practical adaptations for attacking finite mixtures.}
To generate an adversarial example in practice, one can directly maximize the EOL objective in \eqref{eq:attack:eot-finite-mixture} (as in \eqref{eq:attack-finite-mixtures-eot-loss}), or maximize the loss computed on the expected output of the mixture (\eqref{eq:attack-finite-mixtures-eot-output}) \cite{pinot2020randomization,tramer2020adaptive}.
\begin{equation} \label{eq:attack-finite-mixtures-eot-loss}
    \displaystyle \sup_{x' \in B_p(x, \epsilon)} \quad \displaystyle\sum_{i = 1}^{m} ~\mixweights^{(i)} ~ \loss(h_i, x', y).
\end{equation}
\begin{equation} \label{eq:attack-finite-mixtures-eot-output}
      \displaystyle \sup_{x' \in B_p(x, \epsilon)}  \quad  \loss \left(z \mapsto \displaystyle\sum_{i = 1}^{m} ~\mixweights^{(i)} ~ h_i(z), x', y \right).
\end{equation}
In practice, both of these problems are solved using first-order optimization methods like PGD. We refer to these attacks as EOL-PGD and LOE-PGD, respectively. 
The problem in doing so is that there is an underlying assumption that all classifiers can be attacked simultaneously and that their vulnerabilities are \emph{aligned}, a concept we illustrate in Figure \ref{fig:attack:alignment-high-low}.
This is because the gradient of either objective can be rewritten as a linear combination of the gradients of the loss of the individual classifiers with positive coefficients (see Appendix \ref{sec:app:grad-computations} for more details).
Thus, using first-order methods to solve problems \eqref{eq:attack-finite-mixtures-eot-loss} or \eqref{eq:attack-finite-mixtures-eot-output} is intuitively trying to attack all classifiers simultaneously, and the success of such attack relies on the assumption that this linear combination is a good attack direction. This can be effective in a scenario like the one depicted in Figure \ref{fig:attack:alignment-high-low} (left), in which all classifiers are vulnerable and their vulnerabilities are aligned, but fail when the vulnerabilities lie outside the set of admissible perturbations, as demonstrated in Figure \ref{fig:attack:alignment-high-low} (right). 

This issue with EOL-PGD and LOE-PGD is the starting point for the development of ARC~\cite{dbouk2022adversarial}, a stronger attack against mixtures of classifiers, which we will discuss in the following section.

\subsection{ARC attack and its limitations}

Dbouk et al.~\cite{dbouk2022adversarial} criticize EOL-PGD and LOE-PGD because of their lack of \emph{consistency}\footnote{This is the term used by authors in the original work. We rather use the term \emph{effectiveness}. See Section \ref{sec:attack:desirable-properties}} \cite[Theorem 5.1]{dbouk2022adversarial}, meaning that even if there exist adversarial attacks that increase the error of the mixture, EOL-PGD or LOE-PGD can miss them (\emph{i.e.} Figure \ref{fig:attack:alignment-high-low} (right)). This motivates the authors to create \textit{Attacking Randomized ensembles of Classifiers} (ARC)\cite{dbouk2022adversarial}, an attack against finite mixtures of classifiers that is guaranteed to be consistent in binary classification against linear classifiers. In the rest of this section, we will compare ARC to EOL-PGD and discuss the limitations of ARC. We omit the discussion of LOE-PGD, but our analysis applies to it as well.

\begin{figure}[ht]
    \begin{minipage}{0.47\textwidth}
        \centering
        \includegraphics[scale=0.5]{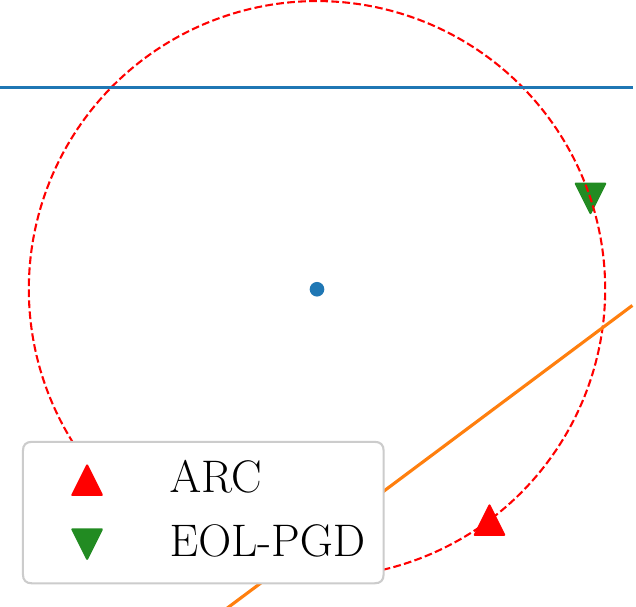}
    \end{minipage}\hfill
    \begin{minipage}{0.47\textwidth}
        \centering
        \includegraphics[scale=0.5]{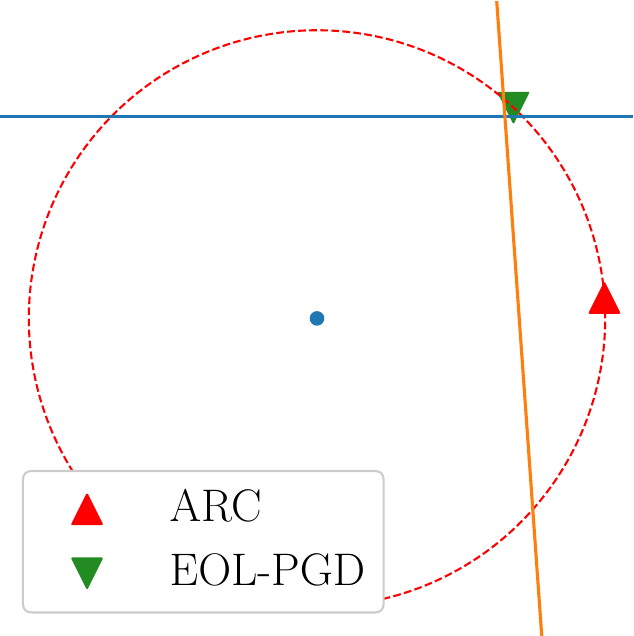}
    \end{minipage}\hfill
    \caption{Comparison between the perturbation proposed by EOL-PGD and ARC on a mixture of two linear classifiers with low to medium alignment.}
    \label{fig:attack:apgd-problem-arc-solution-problem}
\end{figure}

The key difference between ARC and EOL-PGD is that ARC attacks the classifiers one by one, instead of trying to attack all of them simultaneously. At each iteration, ARC follows the direction that attacks the one classifier at hand, and at the end of the iteration, the new perturbation will be kept only if the error of the whole mixture was strictly increased (see \cite[Algorithm 1]{dbouk2022adversarial} for full details).
The greedy approach of ARC makes it provably consistent when attacking binary linear classifiers (see \cite[Theorem 5.1]{dbouk2022adversarial}).
In Figure \ref{fig:attack:apgd-problem-arc-solution-problem} (left) we revisit a situation akin to Figure \ref{fig:attack:alignment-high-low} (right), and we can see that ARC is able to attack one of the classifiers successfully, while EOL-PGD fails to find an adversarial perturbation.


Dbouk et al.~\cite{dbouk2022adversarial} train mixtures of two classifiers using the method proposed in \cite{pinot2020randomization} and find that when attacked with ARC\footnote{ARC is born from an analysis in the binary linear case, but it is generalized to the case of multiclass differentiable classifiers like neural networks. In their experiments comparing to the results in Pinot et al.~\cite{pinot2020randomization}, they used the latter.}, the robustness of these mixtures drops significantly compared to when they are attacked with EOL-PGD or LOE-PGD, which were the attacks used by Pinot et al.~\cite{pinot2020randomization} for their evaluation. This behavior remains consistent across neural network architectures, datasets, and norms considered. Thus, ARC has proven to be a much stronger attack than EOL-PGD and LOE-PGD and remains, to this day, the state-of-the-art attack against finite mixtures of classifiers.

Although ARC resolves the main issue with EOL-PGD, it does so at the cost of losing the ability to attack multiple classifiers simultaneously in situations of medium alignment. Experimentally, one can verify that even for two linear classifiers in $\mathbb{R}^2$, ARC may fail to find a perturbation that is misclassified by both classifiers simultaneously when the region of common vulnerabilities within the $\epsilon$-ball is very small. An example of such scenario is shown in Figure \ref{fig:attack:apgd-problem-arc-solution-problem} (right). 

\begin{figure}[ht]
  \begin{minipage}{0.33\textwidth}
    \centering
    \includegraphics[width=0.95\textwidth]{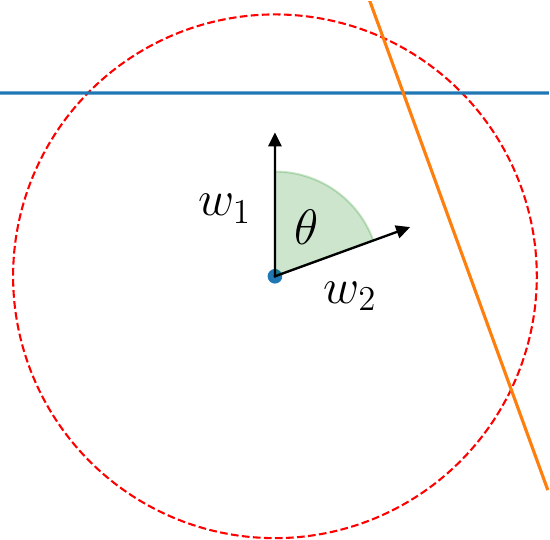}
  \end{minipage}\hfill
  \begin{minipage}{0.65\textwidth}
    \centering
    \includegraphics[width=0.85\textwidth]{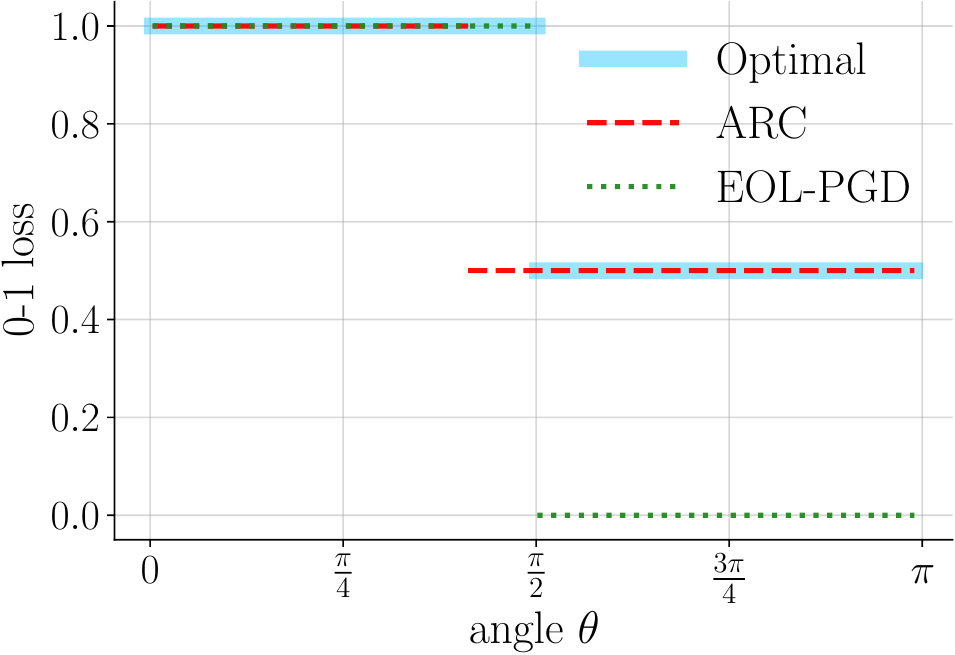}
  \end{minipage}
  \caption{Comparison between EOL-PGD and ARC w.r.t the level of alignment of the linear classifiers. On the left figure, a schema of the experiment that we run for a fixed angle $\theta$. On the right figure, the error of the perturbations found by EOL-PGD and ARC compared to the optimal error w.r.t. $\theta$. 
  }
  \label{fig:attack:apgd-arc-against-optimal}
\end{figure}

To further demonstrate this issue, we test both EOL-PGD and ARC in situations that range from high to low alignment by changing the angle between the normal vectors of the linear classifiers from 0 degrees (perfect alignment) to 180 degrees (opposite normal vectors, low alignment). 
In Figure \ref{fig:attack:apgd-arc-against-optimal}, we plot the error induced by the perturbations found by both EOL-PGD and ARC and compare it to the optimal error, which is determined by checking if the intersection of the two linear classifiers lies within the $\epsilon$-ball or not. It can be seen that no attack dominates the other and that each outperforms the other in some regime that depends on the alignment of the decision boundaries. 

We end this section with a comparison of ARC, EOL-PGD and LOE-PGD through the lens of the directions used to craft the perturbation at each iteration of the attack.
Recall that the gradient directions used by EOL-PGD and LOE-PGD were linear combinations of the gradients of the loss of individual classifiers with non-zero coefficients for \emph{all} of them. In the case of ARC, these coefficients are all zero except for one, which corresponds to the classifier being considered at the current step (See \cite[Algorithm 2]{dbouk2022adversarial} for full details).

\subsection{Desirable properties of an attack for mixtures} \label{sec:attack:desirable-properties}
In this section we will formalize the problem of attacking a mixture of classifiers as a combinatorial optimization problem, specifically as the problem of enumerating maximal elements of a lattice. Using this formalism, we will discuss the desirable properties for attack algorithm. This will allow us to identify a new property unsatisfied by existing algorithms: \emph{maximality}.

\paragraph{Formalizing the problem of attacking a finite mixture.}
Note that a deterministic classifier $h$ partitions the input space into at most $K$ non-overlapping regions $h^{\text{-}1}(j)~\subseteq~\cal X,~j\in~[K]$, where each region corresponds to a class.
Moreover, an optimal untargeted attacker only cares about forcing a prediction \emph{different} from the true label $y$ inside the set of admissible perturbations $B_p(x, \epsilon)$. In that sense, each classifier $h$ induces a partition of the set of admissible perturbation $B_p(x, \epsilon)$ into two disjoint regions of interest:
\begin{equation}\label{eq:space-split}
  B_p(x, \epsilon) =  \underbrace{B_p(x, \epsilon) \cap h^{-1}(y)}_{\text{non-adversarial}} \quad \sqcup \quad {\color{red}\underbrace{B_p(x, \epsilon) \cap \bigcup_{j \ne y} h^{-1}(j)}_{\text{Vulnerability region } \mathrm{V}(h)}}.
\end{equation}
We call \emph{vulnerability region} of $h$ the set $B_p(x, \epsilon) \cap \bigcup_{j \ne y} h^{-1}(j)$, and denote it as $\mathrm{V}(h)$. 
With the definition of vulnerability region, attacking  $h$ is equivalent to finding a point $x'\in \mathrm{V}(h)$, so  the error under attack of $\mixture$ at $(x, y)$ becomes: 
\begin{align} \label{eq:attack:sup-error-mixture-rewritten}
  \zeroOneloss_{\epsilon, p}(\mixture, x, y) & =  \sup_{x' \in B_p(x, \epsilon)} ~\sum_{i=1}^m \mixweights^{(i)} ~ \1{x' \in \mathrm{V}(h_i)}.
\end{align}
With Equation~\eqref{eq:attack:sup-error-mixture-rewritten} it is clear that the optimal attack belongs to the intersection of the vulnerability regions of the classifiers that maximizes the total mass according to $\mixweights$. Let us define the \emph{common vulnerability region} of a subset of classifiers $\mixset' \subseteq \mixset$ as $\mathrm{CV}(\mixset') = \bigcap_{h \in \mixset'} \mathrm{V}(h) \setminus \bigcup_{h \notin \mixset'} \mathrm{V}(h)$.
In simple terms, $\mathrm{CV}(\mixset')$ is the set of points that are adversarial for exactly the classifiers in $\mixset'$, and thus, if $x' \in \mathrm{CV}(\mixset')$, then $\zeroOneloss(\mixture, x', y) = \sum_{i~\mathrm{s.t.}~ h_i \in \mixset'} \mixweights^{(i)}$,
and if $\mixset_1 \subseteq \mixset_2$ and $\mathrm{CV}(\mixset_2) \ne \emptyset$, then the following holds:
\begin{equation}\label{eq:attack:cv-ineq-prefered}
  \forall x_1 \in \mathrm{CV}(\mixset_2),~ \forall x_2 \in \mathrm{CV}(\mixset_1),~\zeroOneloss(\mixture, x_2, y) \le \zeroOneloss(\mixture, x_1, y),
\end{equation}
which means that an attacker prefers to attack in the common vulnerability region of a larger set of classifiers whenever it is not empty.

Equation \eqref{eq:attack:cv-ineq-prefered} suggests that the subsets of classifiers can be ordered by their preference for the attacker. Formally speaking, we can define the partial order
\begin{equation}  \label{eq:attack:order-relation}
  \mixset_1 \preceq \mixset_2  \iff \mixset_1 \subseteq \mixset_2 \text{ and } \mathrm{CV}(\mixset_2) \ne \emptyset.
\end{equation}
The order relation in Equation \eqref{eq:attack:order-relation} induces a \textit{lower semilattice structure} in the family of subsets $\mathcal{S} = \{ \mixset' \subseteq \mixset ~|~ \mathrm{CV}(\mixset') \ne \emptyset \} ~\bigcup~ \{\emptyset\}$.
For simplicity, we refer to it as \emph{adversarial lattice} of $\mixture$ at $(x, y)$. Due to space constraints, we illustrate examples of adversarial lattices in Appendix \ref{app:sup_details}.

The lattice object allows us to discuss desirable properties of an attack algorithm that faces mixtures of classifiers.
The first has already been discussed under the name of consistency \cite{dbouk2022adversarial}, but we redefine it as effectiveness:
\begin{definition}[Effectiveness property] \label{def:effectiveness}
  An attack algorithm is \textit{effective} if for any finite mixture $\mixture$ and point $(x, y)$, it can generate an adversarial example increasing the error of the mixture whenever such a point exists.
\end{definition}

\begin{definition}[Maximality property] \label{def:maximality}
  An attack algorithm is \textit{maximal} if for any finite mixture $\mixture$ and any point $(x, y)$, it can generate an adversarial example for a maximal subset of classifiers of the adversarial lattice.  
\end{definition}

\newcommand{\scaleLattice}{0.7}
\newcommand{\latticeStep}{3}
\begin{figure}[ht]
  \centering
    \begin{tikzpicture}[scale=\scaleLattice,
      every node/.style={draw=black},
      effectivenode/.style={draw=ctcolormain, fill=ctcolormain!20, thick},
      maximalnode/.style={draw=ctcolormain, fill=ctcolormain!20, thick, text=red},
      chosenedge/.style={color=ctcoloraccessory, line width=1.5pt, arrows = {-Latex[length=8pt]}}
      ]
      \node (empty) at (0, 0) {$\emptyset$};
      \node[effectivenode] (a) at (-2*\latticeStep+1, 2) {$\{h_1\}$};
      \node[effectivenode] (b) at (-1*\latticeStep+1, 2) {$\{h_2\}$};
      \node[effectivenode] (c) at (1*\latticeStep-1, 2) {$\{h_3\}$};
      \node[effectivenode] (d) at (2*\latticeStep-1, 2) {$\{h_4\}$};

      \node[effectivenode] (ab) at (-3*\latticeStep+1, 4) {$\{h_1, h_2\}$};
      \node[effectivenode] (ac) at (-2*\latticeStep+1, 4) {$\{h_1, h_3\}$};
      \node[maximalnode] (ad) at (-1*\latticeStep+1, 4) {$\{h_1, h_4\}$};
      \node[maximalnode] (cd) at (3*\latticeStep-1, 4) {$\{h_3, h_4\}$};
      \node[maximalnode] (abc) at (-2*\latticeStep+1, 6) {$\{h_1, h_2, h_3\}$};
  
      \draw (empty) -- (a);
      \draw (empty) -- (b);
      \draw (empty) -- (c);
      \draw (empty) -- (d);
      \draw (a) -- (ab);
      \draw (a) -- (ac);
      \draw (a) -- (ad);
      \draw (b) -- (ab);
      \draw (c) -- (ac);
      \draw (c) -- (cd);
      \draw (d) -- (ad);
      \draw (d) -- (cd);
      \draw (ab) -- (abc);
      \draw (ac) -- (abc);
  \end{tikzpicture}
  \caption{Example of an adversarial lattice for a mixture of four classifiers. 
  }
  \label{fig:lattice-effectiveness-maximality}
\end{figure}
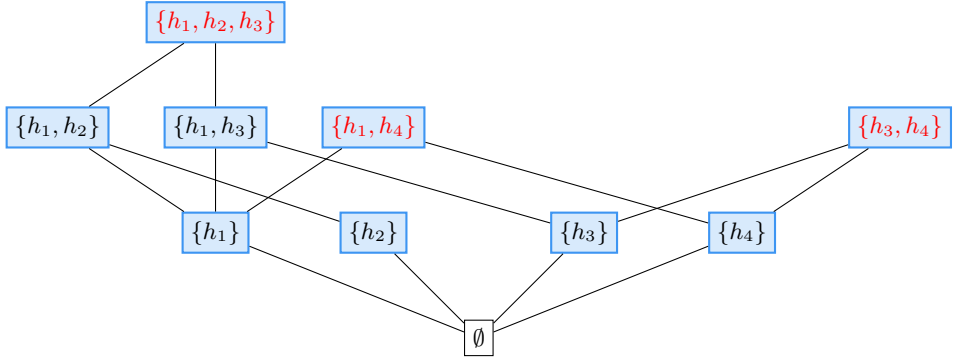

Neither EOL-PGD nor ARC is maximal for binary linear classifiers, as counterexamples can be constructed where they fail to find a maximal subset of classifiers to attack (Figure \ref{fig:attack:apgd-problem-arc-solution-problem}).
Figure \ref{fig:lattice-effectiveness-maximality} illustrates an example of an adversarial lattice for a mixture of four classifiers. An \emph{effective} attack is ensured to produce an adversarial attack $x'$ that corresponds to \emph{some} of the nodes highlighted with a blue rectangle. This property is a defining characteristic of ARC for binary linear classifiers —one that EOL-PGD lacks. However, there is no guarantee regarding the level of the lattice that is reached, and the effectiveness guarantee becomes less meaningful as the size of the lattice increases.
The nodes with red font represent the maximal nodes, illustrating that the maximality guarantee is a stronger property than effectiveness.

The most desirable property of an attack for mixtures is be \emph{optimality}:
\begin{definition}[Optimality property] \label{def:optimality}
  An attack algorithm is \textit{optimal} if for any finite mixture over $\mixture$ and any point $(x, y)$, it can generate an adversarial example that achieves the highest possible \zeroOne loss. 
\end{definition}
 The optimal attack corresponds to a \emph{subset} of the maximal elements of the adversarial lattice, depending on the weights of the mixture.
Unfortunately, no polynomial-time algorithm is guaranteed to achieve optimality, even when using linear classifiers in binary classification, due to the following result (full proof in Appendix \ref{app:proof-hardness-maxfls}.):

\begin{restatable}[Hardness of attacking linear classifiers]{theorem}{hardnessmaxfls}
\label{thm: hardness maxfls} 
Consider a binary classification setting, $\ell_p$ norm with $p > 1$ and $\epsilon$ to be defined. Given a labeled point $(x,y)$, a set of $m$ linear classifiers $x\mapsto\mathds{1}\left\{ \theta_{i}^{\top}x+b_{i}\ge0\right\} $
where $(\theta_{i},b_{i}) \in\mathbb{R}^{d+1}$, a uniform mixture $\mixture$
composed of these linear classifiers
and a value $\beta>0$, there exists a noise budget $\epsilon>0$ such that the problem of checking if there exists $x'\in B_p(x, \epsilon)$ such that $\zeroOneloss\left(\mixture, x' ,y\right)\ge\beta$
is NP-hard.
\end{restatable}


As a consequence of Theorem \ref{thm: hardness maxfls}, the attack algorithm that we will propose will prioritize \emph{maximality} over optimality, which can be guaranteed in the case of binary linear classifiers and provides effectiveness for free.

\section{Lattice Climber attack} \label{sec:attack:lca}

In this section, we introduce a new attack called Lattice Climber Attack (LCA). We first present a version of our attack that has maximality guarantees against mixtures of binary linear classifiers under mild assumptions, and then extend it to multiclass differentiable classifiers. Afterwards, we will present experimental results that compare LCA with EOL-PGD and ARC on synthetic and real-world datasets.

The idea behind  LCA is to arrive at a maximal element of the lattice by climbing one level at a time.
Similar problems have been studied under the name of \textit{most specific sentences} or \textit{maximal frequent itemsets} in the domain of data mining and knowledge discovery, where algorithms like \verb|AllMSS| \cite{gunopulos1997discovering} have been proposed. Our approach is similar to \verb|A_Random_MSS| in \cite{gunopulos1997discovering}.

\subsection{Binary linear classifiers} \label{sec:attack_blc}

In this section and for simplicity, we consider class labels $y \in \lbrace -1, 1 \rbrace$ and classifiers $h : \mathbb R^d \rightarrow \lbrace -1, 1 \rbrace$ of the form $h(x) = \mathrm{sign}(f(x))$ for some linear function $f : \mathbb R^d \rightarrow \mathbb R$. In this setting, $h$ correctly classifies the data point $(x,y)$ if $ y f(x) > 0$. Therefore, attacking $h$ translates to minimizing $y f(x)$. The optimal attack direction and margin to the decision boundary of a single linear classifier are known for all $\ell_p$ norm with $p \ge 1$ \cite[Appendix A]{dbouk2022adversarial}.

The first component of our attack algorithm is a procedure to attack a subset of classifiers $\mixset'$ and find $x + \delta \in \mathrm{CV}(\mixset')$ whenever $\mathrm{CV}(\mixset') \neq \emptyset$. 
Similar to \cite{perdomo2019robust}, we consider the \emph{reverse hinge loss} $\loss^{rev}(y f(x)) = \max \left( y f(x), 0 \right)$ as an objective to \emph{minimize} in order to attack \emph{one} binary linear classifier. 
Attacking a binary linear classifier with the traditional hinge loss would imply maximizing a convex function, for which there is no guarantee of convergence to a global optimum using algorithms like PGD. 
On the other hand, using the reverse hinge loss is equivalent to \emph{minimizing} a bounded convex function, for which PGD is guaranteed to converge to a global optimum.

Another useful property of the reverse hinge loss is that the expected reverse hinge loss over a finite mixture of classifiers is convex and equal to zero if and only if all the classifiers in the mixture misclassify $x$.
Then, in order to attack a set of classifiers $\mixset'$, we can minimize the \emph{sum of reverse hinge losses} \cite{perdomo2019robust}
\begin{equation}
\mathrm{SRH}(\mixset', x, y) =  \frac{1}{\lvert \mixset' \rvert} \sum_{h \in \mixset'} \loss^{rev}(y f(x)).
\end{equation}
If there exists $x'$ such that $\mathrm{SRH}(\mixset', x', y) =  0$, then $\mathrm{PGD}$ with an appropriately chosen step size and sufficient number of iterations will converge to some $x''$ that is misclassified by all classifiers in $\mixset'$ (see \cite[Theorem 3]{perdomo2019robust} for details).

The second component of our attack algorithm is a way to navigate the adversarial lattice. 
We propose a bottom-up navigation mechanism to climb a branch of the adversarial lattice that consists of keeping a pool of fooled classifiers, attempting to add each classifier in a fixed order. A classifier joins the pool only if it can be fooled alongside all current members; otherwise, it is discarded. The algorithm terminates after evaluating all classifiers.

The order in which classifiers are considered is a parameter of the algorithm, and similarly to \cite{dbouk2022adversarial}, we find that considering them in decreasing order of their associated weight yields good performance in general. For example, in the case $m=2$, and with suitable parameters for the internal $\mathrm{PGD}$, it ensures that LCA is optimal. Another reasonable option is to use a random permutation of the classifiers, which Dbouk et al.~\cite{dbouk2023robustness} found to be useful for ARC. The pseudocode of LCA in the binary linear setting is shown in Algorithm \ref{alg:binary_linear_attack}.

\begin{algorithm}[!t]
\caption{LCA for binary linear classifiers}\label{alg:binary_linear_attack}
\begin{algorithmic}[1]
\REQUIRE Set $\mixset$ of $m$ binary linear classifiers in some order $(h_1, \dots, h_m)$, starting point $(x, y) \in \mathbb{R}^d \times \{-1 , 1\}$. $T$ number of iterations and $\eta$ step size for $\mathrm{PGD}$.
\STATE Initialize pool $\mixset_{\text{pool}} = \emptyset$, $\delta = 0_{d}$
\FOR{$ i = 1, 2 \cdots, m$}\label{algo:blc:forloop}
 \STATE $\mixset_{\text{pool}} = \mixset_{\text{pool}} \cup \{ h_i \}$  \Comment{Add $h_i$ to the pool} \label{algo:blc:add_to_pool}
 \STATE Attack $\mathrm{SRH}(\mixset_{\text{pool}}, \cdot, y)$ starting at $x + \delta$ with $\mathrm{PGD} \left( T, \eta \right)$ to find new perturbation $\hat{\delta}$  \label{algo:blc:attack_current_pool}
 \IF{$\mathrm{SRH}(\mixset_{\text{pool}}, x + \hat{\delta}, y) = 0$ \textit{i.e.} succeeded}
    \STATE $\delta = \hat{\delta}$ \Comment{Update current attack, keep $h_i$ in pool} \label{algo:blc:update_point}
 \ELSE{} 
    \STATE $\mixset_{\text{pool}} = \mixset_{\text{pool}} \setminus \{ h_i \}$ \Comment{Reset pool to last state}\label{algo:blc:reset_pool}
\ENDIF
\ENDFOR
\STATE \textbf{return} Adversarial example $x + \delta$\label{algo:blc:return-end}
\end{algorithmic}
\label{algo_blc}
\end{algorithm}
Recall that the step used by EOL-PGD can be expressed as linear combinations of the gradients of the loss of individual classifiers. For LCA, similar to ARC, the coefficients of this linear combination will be sparse: only the models within the pool $\mixset_{\mathrm{pool}}$ will have non-zero coefficients. This behavior positions LCA as an intermediate approach between EOL-PGD and ARC: in LCA, the coefficients initially resemble those in ARC, but as the pool grows, they transition to resemble the denser pattern of EOL-PGD. 
This behavior makes it possible to adapt to different types of adversarial lattice: when classifiers are not simultaneously vulnerable, LCA will behave like ARC, which was created to ensure \emph{effectiveness} for binary linear classifiers, and when they are simultaneously vulnerable, it will behave like EOL-PGD which attacks all classifiers simultaneously.

\paragraph{Maximality of LCA in the binary linear setting.}
In the binary linear classifier setting, $\mathrm{PGD}(T, \eta)$ is guaranteed to minimize the function $\mathrm{SRH}(\mixset_{\text{pool}}, \cdot, y)$ because it is a convex optimization problem \cite{perdomo2019robust}. Thus, line \ref{algo:blc:attack_current_pool} of Algorithm \ref{alg:binary_linear_attack} ensures that we will find an adversarial attack $x + \hat{\delta} \in \mathrm{CV}(\mixset_{\text{pool}})$ if $\mathrm{CV}(\mixset_{\text{pool}}) \neq \emptyset$. This enables climbing to the top of the adversarial lattice along a branch determined by the classifier order.
This is formalized in the following lemma:
\begin{restatable}{lemma}{lemmaLcaMaximal} %
    \label{lemma:climb_step}
    Consider a set of $m$ binary linear classifiers $\mixset$. Fix $\epsilon > 0$ the attack budget and a point $(x, y)$. If there exists a point $x' \in B_{p}(x, \epsilon)$ such that $\mathrm{SRH}(\mixset, x', y)=0$, then there exist parameters $T$ and $\eta$ such that minimizing $\mathrm{SRH}(\mixset, x, y)$ w.r.t. $x$ with $\mathrm{PGD}(T, \eta)$ will return $x''$ such that $\mathrm{SRH}(\mixset, x'', y)=0$.
\end{restatable}
The proof of Lemma \ref{lemma:climb_step} is given in Appendix \ref{app:proof-climb-step}. Lemma \ref{lemma:climb_step} allows us to prove that LCA is maximal for the set of binary linear classifiers (full proof in Appendix \ref{app:proof-binary-maximality}):
\begin{restatable}[LCA is maximal in the binary linear setting]{theorem}{thmLcaMaximality}
    \label{thm:binary_maximality}
    Let $\mixset$ be a finite set of binary linear classifiers and $\mixture$ a mixture over $\mixset$. Fix $\epsilon > 0$ the attack budget and a $p$-norm with $p>1$. For any $(x, y)~\in~\mathbb{R}^d~\times~\{-1,1\}$, there exist parameters $T$ and $\eta$ for the inner $\mathrm{PGD}$ such that Algorithm \ref{alg:binary_linear_attack} returns an adversarial example $x + \delta \in \mathrm{CV}(\mixset')$, where $\mixset'$ is a maximal element of the adversarial lattice of $\mixture$ at $(x, y)$.
\end{restatable}

\subsection{Multiclass differentiable classifiers} \label{sec:attack_multi}

The ideas developed in Section \ref{sec:attack_blc} for binary linear classifiers need to be adapted to the multiclass case with general differentiable classifiers, like neural networks, because we cannot have the guarantees provided by Lemma~\ref{lemma:climb_step} and Theorem~\ref{thm:binary_maximality}. Moreover, the reverse hinge loss was defined for binary classifiers and not for multiclass classifiers.

In order to attack a classifier $h$ that predicts the class with the highest score according to the score function $f: \cal X \to \mathbb{R}^K$ at an arbitrary point $(x, y)$, we choose the target label $y_{\mathrm{adv}} \in [K] \setminus \{y\}$ with the largest score according to $f(x)$ and minimize the reverse hinge loss of the margin between $f(x)^{(y)}$ and $f(x)^{(y_{\mathrm{adv}})}$, \textit{i.e.} $\loss^{rev}(f(x)^{(y)} - f(x)^{(y_{\mathrm{adv}})}).$
This is a common choice that has been used by Perdomo et al.~\cite{perdomo2019robust} and also by Carlini et al.~\cite{carlini2017towards}, who found it to perform well as an objective function for crafting adversarial attacks.
Our proposed attack is presented in Algorithm \ref{alg:multiclass_attack} in Appendix \ref{app:lca:algo:multi}. As there is no guarantee on the convergence of the attack to $\mathrm{SRH}(\mixset_{\text{pool}}, \cdot, y)$ in line \ref{algo:attack_current}, we change the criteria to update the pool of classifiers: each time we find a perturbation $\delta$ with a higher error for the mixture, we keep it and update the pool to be the classifiers that misclassify the current adversarial example $x +\delta$.

\section{Experiments} \label{sec:attack:xp}

In this section, we show the experimental results that support the theoretical guarantees of LCA in the binary linear setting. We also compare the performance of LCA with ARC and EOL-PGD in the multiclass differentiable setting, in which guarantees are not provided, and show that LCA generally performs better than existing state-of-the-art attacks. More details on the experimental setup and additional experiments can be found in Appendix \ref{app:toy} and \ref{app:cifar}.

\subsection{Synthetic data: Linear classifiers} \label{sec:experiments:synthetic}

\paragraph{Binary linear classifiers in high dimension.} In this experiment, we assess how the performance of EOL-PGD, ARC and LCA scale with the number of classifiers $m$ in a higher dimension $d$. For each value $m$ and a fixed value of $\epsilon=1$, we repeat the following experiment 1000 times: we fix our point $(x,y) = (0_d, -1)$ and sample $m$ i.i.d linear classifiers $(w_i, b_i)$ where $w_i$ is uniformly sampled from the unit sphere in $\mathbb{R}^d$ and $b_i \sim -|\mathcal{N}(\alpha, \beta)|$ follows a folded Gaussian distribution, and we test the three attacks against the sampled mixture. The hyperparameter $\alpha$ controls the expected distance from $x$ to the decision boundary of the classifiers, and $\beta$ the variance of such distance.

Figure \ref{fig: 2d score num models plot} shows that the average performance of all attacks deteriorates as the number of models increases, but  LCA remains superior for all $m$. Note that depending on the difficulty of the configurations ($\alpha$ and $\beta$), the performance of the attacks can change drastically. Nevertheless, LCA remained superior in all configurations, regardless of the dimension. More combinations of $\alpha$ and $\beta$ and more details on the experimental setup can be found in Appendix \ref{app:toy}. 

\begin{figure}[ht]
    \begin{minipage}{0.5\textwidth}
      \centering
      \includegraphics[width=\textwidth]{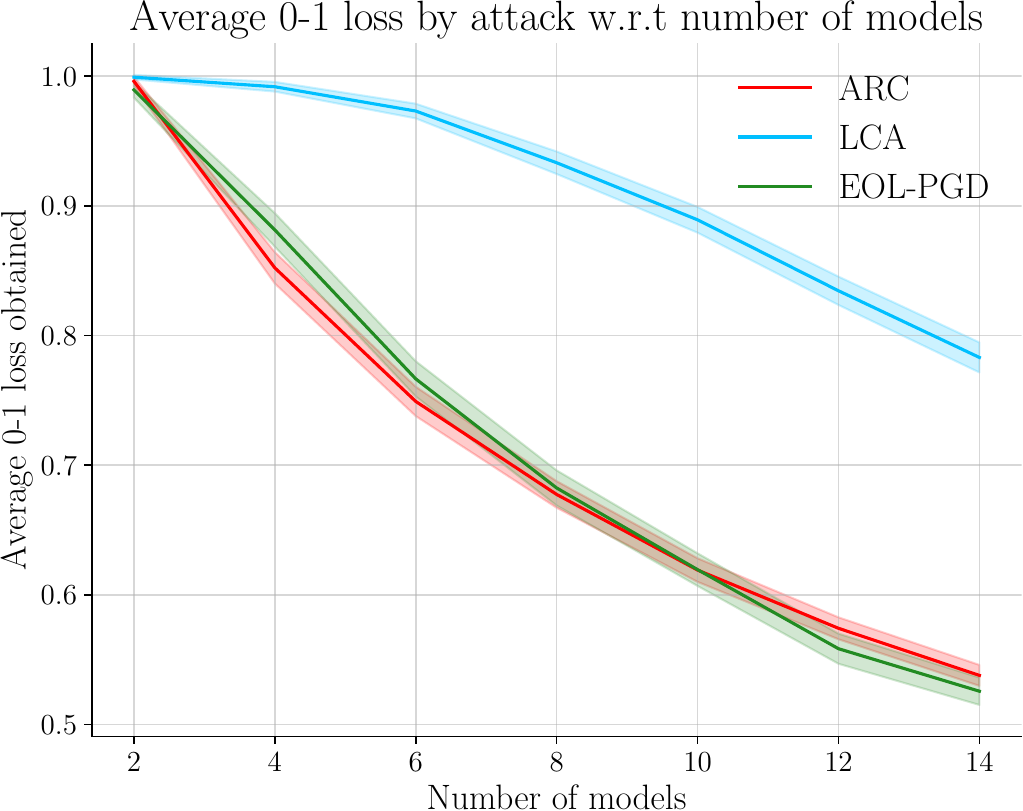} 
    \end{minipage}\hfill
    \begin{minipage}{0.5\textwidth}
      \centering
      \includegraphics[width=\textwidth]{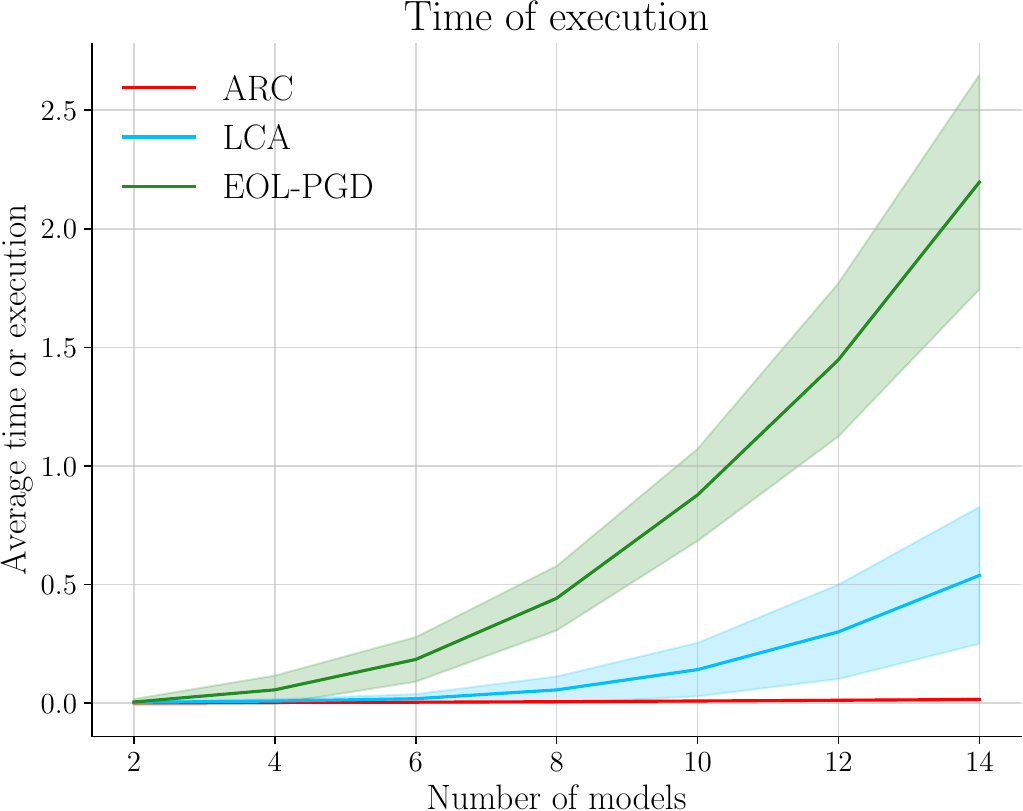} 
    \end{minipage}
    \captionsetup{skip=8pt}
      \caption{On the left figure, the average error obtained by the attacks in $\mathbb{R}^{128}$ as a function of the number of classifiers in the mixture with 99\% confidence intervals. The right figure shows the average time in seconds that each attack took to generate an adversarial example as a function of the number of classifiers with one standard deviation intervals.
      For each number of models, the experiment was repeated 1000 times. Note that the $y$-axis limits have been adapted for visualization purposes. In this case, $\alpha=0.25$, $\beta=0.2$. } 
      \label{fig: 2d score num models plot}
\end{figure}

\subsection{CIFAR-10}

\renewcommand{\arraystretch}{1.5}
\newcommand{\cwidth}{1.1cm}
\newcolumntype{x}[1]{>{\centering\let\newline\\\arraybackslash\hspace{0pt}}p{#1}}
\begin{table*}[t]
\setlength{\aboverulesep}{0pt}
\setlength{\belowrulesep}{0pt}
    \caption{Expected accuracy of mixtures against EOL-PGD, ARC and  LCA under the $\ell_{\infty}$ threat model. Attacker wants to minimize accuracy, so lower is better. All models use ResNet20 as the base architecture, except $\dag$ which uses ResNet18.}
      \centering
      \scriptsize
            \begin{tabular}{|x{1.3cm}x{1.4cm}|x{\cwidth}|x{\cwidth}x{\cwidth}x{\cwidth}|x{\cwidth}x{\cwidth}x{\cwidth}|}
            \toprule
            \multicolumn{3}{|c|}{} & \multicolumn{3}{c|}{$ \epsilon=0.01$} & \multicolumn{3}{c|}{$ \epsilon=0.03  $} \\
            \midrule
            \multicolumn{2}{|c|}{Model} &  Nat & EOL & ARC &  LCA & EOL & ARC &  LCA \\
            \midrule
            BASE/3 &                            & 91.8\% & 15.0\% & 4.3\% & \textbf{1.7\%} & 0.1\% & 0.2\% & \textbf{0.0\%} \\ 
            BASE/5 &                            & 91.9\% & 12.3\% & 6.1\% & \textbf{2.7\%} & \textbf{0.0\%} & 0.2\% & 0.1\% \\ 
            BASE/8 &                            & 91.8\% & 10.6\% & 7.9\% & \textbf{4.6\%} & \textbf{0.0\%} & 0.2\% & 0.1\% \\
            \midrule
            ADP/3 &                             & 89.4\% & 15.7\% & \textbf{14.6\%} & 15.1\% & \textbf{0.4\%} & 5.2\% & 1.8\% \\ 
            ADP/5 & \cite{pang2019adp}          & 88.8\% & 15.4\% & 16.1\% & \textbf{13.9\%} & \textbf{0.2\%} & 2.5\% & 1.5\% \\ 
            ADP/8 &                             & 88.5\% & \textbf{15.2\%} & 18.4\% & 15.4\% & \textbf{0.3\%} & 4.2\% & 3.3\% \\ 
            \midrule
            GAL/3 &                             & 87.1\% & 43.0\% & 18.0\% & \textbf{14.3\%} & 14.5\% & 5.8\% & \textbf{0.9\%} \\ 
            GAL/5 & \cite{kariyappa2019gal}     & 88.7\% & 46.2\% & 45.1\% & \textbf{36.0\%} & 11.5\% & 9.4\% & \textbf{7.1\%} \\ 
            GAL/8 &                             & 89.7\% & \textbf{44.5\%} & 52.4\% & 50.2\% & \textbf{3.9\%}& 11.7\% & 15.9\% \\ 
            \midrule
            AT/3 &                              & 77.3\% & 68.6\% & 68.3\% & \textbf{65.3\%} & 47.2\% & 47.7\% & \textbf{42.4\%} \\ 
            AT/5 & \cite{madry2017towards}      & 77.9\% & 69.0\% & 69.1\% & \textbf{65.2\%} & 47.2\% & 49.2\% & \textbf{42.7\%} \\ 
            AT/8 &                              & 77.8\% & 69.0\% & 69.0\% & \textbf{64.9\%} & 47.5\% & 50.3\% & \textbf{44.3\%} \\ 
            \midrule
            DV/3 &                              & 89.8\% & 52.6\% & 44.4\% & \textbf{31.2\%} & 39.2\% & 5.8\% & \textbf{1.9\%} \\ 
            DV/5 & \cite{yang2020dverge}        & 89.9\% & 59.0\% & 55.1\% & \textbf{42.1\%} & 35.3\% & 10.1\% & \textbf{6.0\%} \\ 
            DV/8 &                              & 88.6\% & 62.1\% & 63.2\% & \textbf{52.8\%} & 31.7\% & 18.4\% & \textbf{15.0\%} \\ 
            \midrule
            DV+AT/3 &                           & 81.7\% & 71.4\% & 71.3\% & \textbf{69.0\%} & 44.4\% & 45.2\% & \textbf{41.0\%} \\ 
            DV+AT/5 & \cite{yang2020dverge}     & 84.0\% & 72.8\% & 73.1\% & \textbf{70.2\%} & 42.2\% & 44.6\% & \textbf{39.9\%} \\ 
            DV+AT/8 &                           & 84.0\% & 73.2\% & 73.4\% & \textbf{70.9\%} & 44.3\% & 46.5\% & \textbf{42.5\%} \\ 
            \midrule
            BARRE/5 & \cite{dbouk2023robustness}& 76.3\% & 68.9\% & 69.4\% & \textbf{66.2\% }& 50.6\% & 49.9\% & \textbf{46.1\%} \\ 
            \midrule
            MR/5 & \cite{zhang2022marginboosting}& 73.7\% & 65.7\% & 66.4\% & \textbf{62.3\% }& 46.3\% & 46.7\% & \textbf{41.5\% }\\ 
            MR/5$^{\dag}$ & & 81.3\% & 72.7\% &74.1\% & \textbf{70.2\% }& 47.1\% & 51.4\% & \textbf{46.5\%} \\ 
            \bottomrule
           \end{tabular}
           \label{table: cifar-10 results}
\end{table*}

\paragraph{Experimental setup.} To test our attack against differentiable multiclass classifiers, we measure the accuracy of mixtures built using models trained with different ensemble diversity techniques. To have a wide variety of models and training methods, we take the pre-trained models from the DVERGE \cite{yang2020dverge} repository\footnote{\url{https://github.com/zjysteven/DVERGE/tree/main}}, which compares baseline models trained without any defense (BASE) against ensemble diversity defenses such as ADP \cite{pang2019adp}, GAL \cite{kariyappa2019gal}, plain adversarial training (AT) \cite{madry2017towards} and DVERGE (DV) \cite{yang2020dverge}. The authors made available 3 independent runs of each method (except for BASE), and we report the average accuracy over these 3 independent runs. We also trained a mixture of 5 models using the MRBoost \cite{zhang2022marginboosting} framework, using the ResNet20 and the ResNet18 architectures. All these methods are designed to promote ensemble diversity and reduce the joint adversarial vulnerability of the sub-models, so we consider them appropriate to evaluate the performance of attacks against mixtures. We also take pre-trained models from the BARRE \cite{dbouk2023robustness} repository\footnote{\url{https://github.com/hsndbk4/BARRE/tree/main}}, in which authors propose a boosting algorithm to specifically train robust mixtures. All these models use the ResNet20 architecture as the base classifiers, except for MRBoost with ResNet18. We use the $\ell_{\infty}$ threat model with $\epsilon = 0.03$ as in standard practice, and also the $\epsilon=0.01$ setting to compare with \cite{yang2020dverge}. More details in Appendix \ref{app:cifar}.

To provide a fair comparison, we adjusted the number of iterations $T$ for each attack to ensure that  LCA was not given any advantage. Note that the number of gradient computations is $T \cdot m$ for EOL-PGD, at most 4 $T \cdot m$ for ARC \cite{dbouk2022adversarial}, and at most $T \cdot \frac{m(m+1)}{2}$ for LCA. In our case $m \in \{3, 5, 8\}$. Taking this into account, we set the number of iterations to 50 for LCA, and gave ARC and EOL-PGD up to 200 and 500 iterations respectively. We also tested ARC and EOL-PGD with fewer iterations, and report only the best result. We further give more advantage to EOL-PGD by allowing it to perform random initialization and 5 restarts. In contrast, ARC and LCA do not use random initialization or restarts.

\paragraph{Results.} Table \ref{table: cifar-10 results} shows the results of our evaluation. First, we can confirm the observations made in \cite{dbouk2022adversarial}: EOL-PGD, even with the advantage given, tends to overestimate the robustness of mixtures, and it is more notorious for BASE models with $\epsilon=0.01$ and DV models. In these cases, ARC dramatically outperforms EOL-PGD (as was seen in \cite{dbouk2022adversarial}), and \textit{ LCA outperforms both ARC and EOL-PGD}.
Note however that EOL-PGD is better than both ARC and  LCA when models are simultaneously vulnerable and the individual gradients used for the attack are highly aligned \cite{dbouk2022adversarial,kariyappa2019gal}, which seems to be the case for ADP models. This could be explained by the fact that ADP, contrary to methods like GAL, does not explicitly reduce gradient alignment during training. 

GAL models show a peculiar behavior: robustness against EOL-PGD tends to \textit{decrease} as the number of models increases, suggesting that the models become more aligned. A similar behavior was also reported in \cite[Table 4]{yang2020dverge}, suggesting that this method does not scale well with the number of models. However, robustness against ARC or LCA increases with the number of models. This contrast might suggest that GAL models are indeed more diverse locally, but still simultaneously vulnerable inside the $\epsilon$-ball so that EOL-PGD is able to exploit this vulnerability by naively attacking all classifiers enough times.

For the more robust models (AT, DV+AT, BARRE and MR),  LCA shows a clear improvement in performance when compared to both ARC and EOL-PGD. The average gap between LCA and ARC in the $\epsilon = 0.03$ setting is $4.9\%$, with a minimum gap of $3.8\%$.

Due to space constraints, we provide in Appendix \ref{app:cifar} extra experiments that test stronger versions of ARC and LCA, discuss the effect of using random orderings and restars for LCA, and more importantly the effect of using ARC or LCA in adversarial training. We observed that the use of ARC or LCA in adversarial training was counterproductive and that EOL-PGD produced mixtures with better robustness. This behavior was also observed in \cite{dbouk2023robustness}, which led the authors to use EOL-PGD in the adversarial training part of BARRE, instead of ARC.

\section{Conclusions and future work}

In this paper, we explore the properties of attacks for randomized mixtures and introduce the LCA algorithm that is \textit{maximal} in the binary linear setting and demonstrates good empirical performances against multiclass differentiable classifiers.
Despite employing state-of-the-art diversity inducing techniques, our robustness evaluation confirms that the robustness of mixtures largely depends on the robustness of individual models, often matching that of a single model (the robustness of a 5-ResNet20 mixture in \cite{dbouk2023robustness} is similar to the robustness of a single ResNet20 reported in \cite{dbouk2022adversarial} by the same authors), limiting the practical impact of mixtures. This discrepancy between the theoretical advantages of mixtures and their practical robustness underscores the persistent challenge of training robust mixtures. 
Moreover, adversarial training has limited effectiveness for mixtures, as training with stronger adaptive attacks does not produce more robust models (see Appendix \ref{app:adversarial-training-mixtures}). New theoretical and practical approaches are needed to improve the training of robust mixtures and fully leverage their potential, a direction that has been started in \cite{dbouk2023robustness}.

\begin{credits}
\subsubsection{\ackname} This work was carried out using HPC resources from GENCI– IDRIS (AD011014207 and AD011014207R1), and funded by the French National Research Agency (DELCO ANR-19-CE23-0016). Thanks to the reviewers for the valuable feedback.

\subsubsection{\discintname}
The authors have no competing interests to declare that are
relevant to the content of this article.
\end{credits}

%
\bibliographystyle{splncs04}


\begin{thebibliography}{10}
\providecommand{\url}[1]{\texttt{#1}}
\providecommand{\urlprefix}{URL }
\providecommand{\doi}[1]{https://doi.org/#1}

\bibitem{amaldi1995complexity}
Amaldi, E., Kann, V.: The complexity and approximability of finding maximum feasible subsystems of linear relations. Theoretical computer science  \textbf{147}(1-2),  181--210 (1995)

\bibitem{athalye2018obfuscated}
Athalye, A., Carlini, N., Wagner, D.: Obfuscated gradients give a false sense of security: Circumventing defenses to adversarial examples. In: ICML. pp. 274--283. PMLR (2018)

\bibitem{athalye2018synthesizing}
Athalye, A., Engstrom, L., Ilyas, A., Kwok, K.: Synthesizing robust adversarial examples. In: ICML. pp. 284--293. PMLR (2018)

\bibitem{bubeck2019adversarial}
Bubeck, S., Lee, Y.T., Price, E., Razenshteyn, I.: Adversarial examples from computational constraints. In: ICML. pp. 831--840. PMLR (2019)

\bibitem{carlini2017towards}
Carlini, N., Wagner, D.A.: Towards evaluating the robustness of neural networks. In: Symposium on Security and Privacy. pp. 39--57. {IEEE} Computer Society (2017)

\bibitem{croce2020robustbench}
Croce, F., Andriushchenko, M., Sehwag, V., Debenedetti, E., Flammarion, N., Chiang, M., Mittal, P., Hein, M.: Robustbench: a standardized adversarial robustness benchmark. arXiv:2010.09670  (2020)

\bibitem{dbouk2022adversarial}
Dbouk, H., Shanbhag, N.: Adversarial vulnerability of randomized ensembles. In: ICML. pp. 4890--4917. PMLR (2022)

\bibitem{dbouk2023robustness}
Dbouk, H., Shanbhag, N.R.: On the robustness of randomized ensembles to adversarial perturbations. arXiv:2302.01375  (2023)

\bibitem{dhillon2018stochastic}
Dhillon, G.S., Azizzadenesheli, K., Lipton, Z.C., Bernstein, J., Kossaifi, J., Khanna, A., Anandkumar, A.: Stochastic activation pruning for robust adversarial defense. arXiv:1803.01442  (2018)

\bibitem{goodfellow2014explaining}
Goodfellow, I.J., Shlens, J., Szegedy, C.: Explaining and harnessing adversarial examples. arXiv:1412.6572  (2014)

\bibitem{gunopulos1997discovering}
Gunopulos, D., Mannila, H., Saluja, S.: Discovering all most specific sentences by randomized algorithms extended abstract. In: Int. Conf. on Database Theory. pp. 215--229. Springer (1997)

\bibitem{he2019parametric}
He, Z., Rakin, A.S., Fan, D.: Parametric noise injection: Trainable randomness to improve deep neural network robustness against adversarial attack. In: CCVPR. pp. 588--597 (2019)

\bibitem{kariyappa2019gal}
Kariyappa, S., Qureshi, M.K.: Improving adversarial robustness of ensembles with diversity training. CoRR  \textbf{abs/1901.09981} (2019)

\bibitem{madry2017towards}
Madry, A., Makelov, A., Schmidt, L., Tsipras, D., Vladu, A.: Towards deep learning models resistant to adversarial attacks. arXiv:1706.06083  (2017)

\bibitem{meunier2021mixed}
Meunier, L., Scetbon, M., Pinot, R.B., Atif, J., Chevaleyre, Y.: Mixed nash equilibria in the adversarial examples game. In: ICML. pp. 7677--7687. PMLR (2021)

\bibitem{pang2019adp}
Pang, T., Xu, K., Du, C., Chen, N., Zhu, J.: Improving adversarial robustness via promoting ensemble diversity. In: ICML. pp. 4970--4979. PMLR (2019)

\bibitem{panousis2021stochastic}
Panousis, K.P., Chatzis, S., Theodoridis, S.: Stochastic local winner-takes-all networks enable profound adversarial robustness. arXiv:2112.02671  (2021)

\bibitem{perdomo2019robust}
Perdomo, J.C., Singer, Y.: Robust attacks against multiple classifiers. arXiv:1906.02816  (2019)

\bibitem{pinot2020randomization}
Pinot, R., Ettedgui, R., Rizk, G., Chevaleyre, Y., Atif, J.: Randomization matters how to defend against strong adversarial attacks. In: ICML. pp. 7717--7727. PMLR (2020)

\bibitem{raff2019barrage}
Raff, E., Sylvester, J., Forsyth, S., McLean, M.: Barrage of random transforms for adversarially robust defense. In: CCVPR. pp. 6528--6537 (2019)

\bibitem{sitawarin2022demystifying}
Sitawarin, C., Golan-Strieb, Z.J., Wagner, D.: Demystifying the adversarial robustness of random transformation defenses. In: ICML. pp. 20232--20252. PMLR (2022)

\bibitem{tramer2020adaptive}
Tramer, F., Carlini, N., Brendel, W., Madry, A.: On adaptive attacks to adversarial example defenses. Neurips  \textbf{33},  1633--1645 (2020)

\bibitem{xie2017mitigating}
Xie, C., Wang, J., Zhang, Z., Ren, Z., Yuille, A.L.: Mitigating adversarial effects through randomization. In: ICLR (2018)

\bibitem{yang2020dverge}
Yang, H., Zhang, J., Dong, H., Inkawhich, N., Gardner, A., Touchet, A., Wilkes, W., Berry, H., Li, H.: Dverge: diversifying vulnerabilities for enhanced robust generation of ensembles. Neurips  \textbf{33},  5505--5515 (2020)

\bibitem{zhang2022marginboosting}
Zhang, D., Zhang, H., Courville, A., Bengio, Y., Ravikumar, P., Suggala, A.S.: Building robust ensembles via margin boosting. In: ICML. vol.~162, pp. 26669--26692. PMLR (06 2022)

\end{thebibliography}

\newpage
\appendix

\noindent
\textbf{\Large Supplementary Material}

\section{Supplementary details} \label{app:sup_details}
\subsection{Computing the gradients of the EOL and LOE objectives} \label{sec:app:grad-computations}
Here we consider a differentiable surrogate loss function $\loss$ like the cross-entropy loss. As in standard practice, we consider deterministic classifiers $h$ built from score functions $f: \cal X \to \mathbb{R}^K$ using the \texttt{argmax} operator. We abuse notation and write $\loss(h, x, y)$ even if the loss depends on the underlying score function $f$.
First, the gradient of EOL in \eqref{eq:attack-finite-mixtures-eot-loss} can be directly computed due to the linearity of the gradient operator:
\begin{equation} \label{eq:attack:gradient-eot}
  \nabla_x \displaystyle\sum_{i = 1}^{m} ~\mixweights^{(i)} ~ \loss(h_i, x, y) = \displaystyle\sum_{i = 1}^{m} ~\mixweights^{(i)} ~ \nabla_{x} \loss (h_i, x, y).
\end{equation}
For the LOE objective, we consider the classifier $h_{\text{avg}}$ built from the average of the scores of the classifiers in the mixture $f_{\text{avg}}(x) = \sum_{i = 1}^{m} ~\mixweights^{(i)} ~ h_i(x)$.
The gradient of LOE in \eqref{eq:attack-finite-mixtures-eot-output} is equal to:
\begin{equation} \label{eq:attack:gradient-apgd}
  \nabla_x \loss \left(h_{\text{avg}}, x , y \right) =  \displaystyle\sum_{i = 1}^{m} ~\gamma_i ~ \nabla_{x} \loss (h_i, x, y),
\end{equation}
where the coefficients $\gamma_i$ depend on the derivative of $\loss$ w.r.t. $x$ and are given by
\begin{equation}
  \gamma_i = \displaystyle\frac{\mixweights^{(i)} \cdot \loss'(h_{\text{avg}}, x, y)}{\loss'(h_i, x, y)}.
\end{equation}
\newpage

\subsection{Example of adversarial lattice and different configurations}
\def\figscale{0.15}
\begin{figure*}[!ht]
    \centering
    \captionsetup{width=.25\linewidth,labelformat=empty}
    \minipage{0.25\linewidth}
    \centering
      {
        \includegraphics[scale=\figscale]{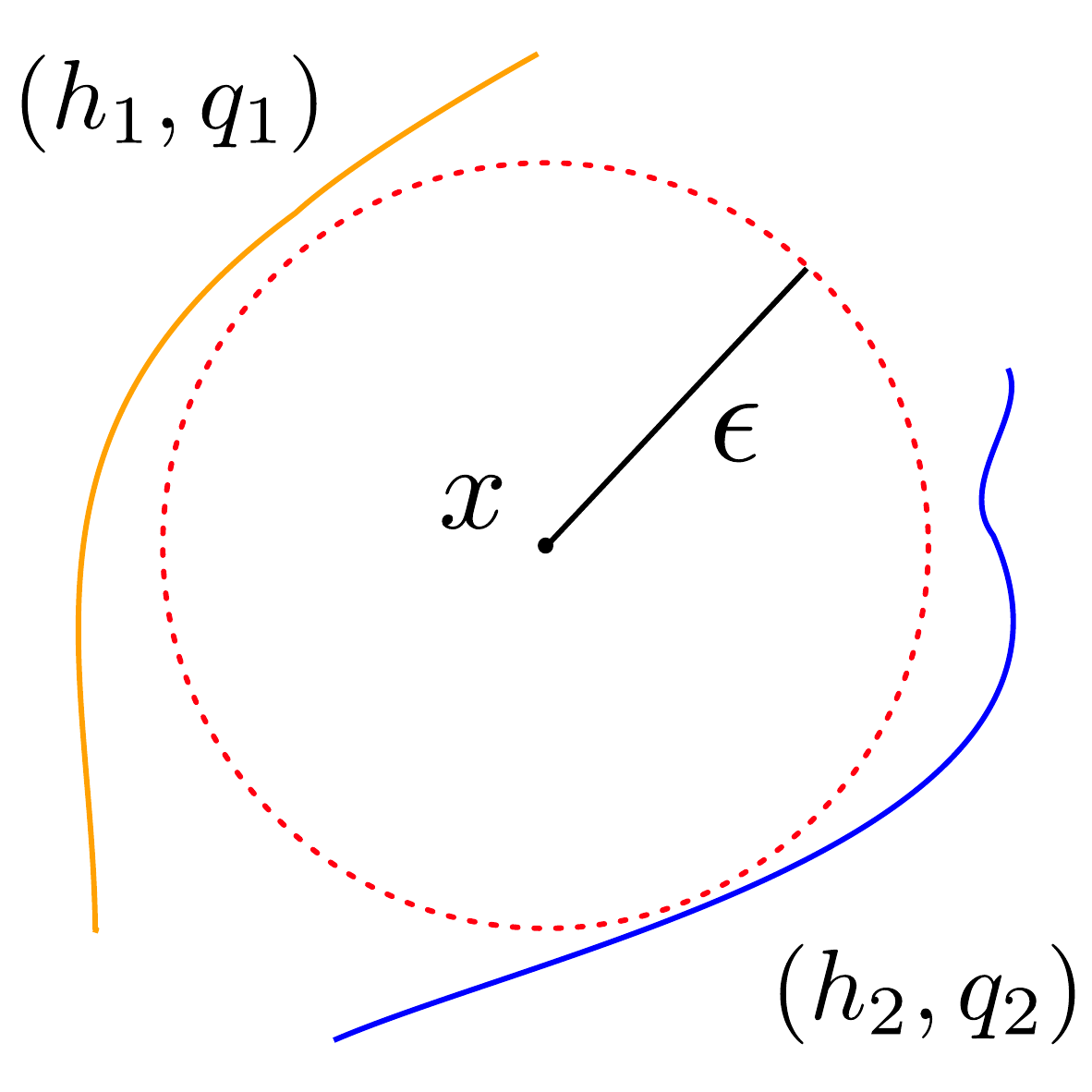}
        \caption{(a)}
        \label{fig:mixtures-lca-two_classifiers_regions_rob}
      }
    \endminipage\hfill
    \minipage{0.25\linewidth}%
    \centering
      {
        \includegraphics[scale=\figscale]{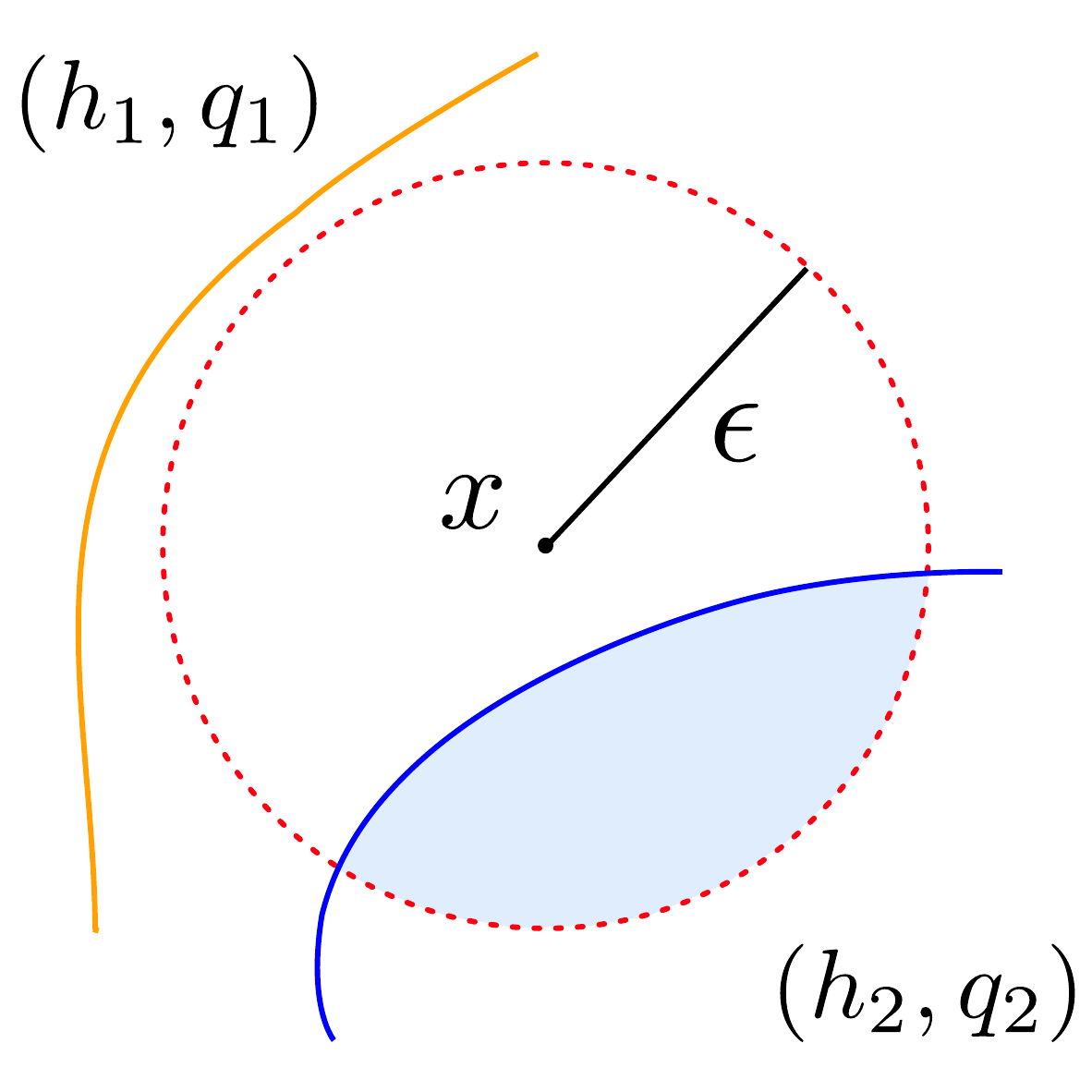} 
        \caption{(b)}
        \label{fig:mixtures-lca-two_classifiers_regions_onerob}
      }
    \endminipage\hfill
    \minipage{0.25\linewidth}
    \centering
      {
        \includegraphics[scale=\figscale]{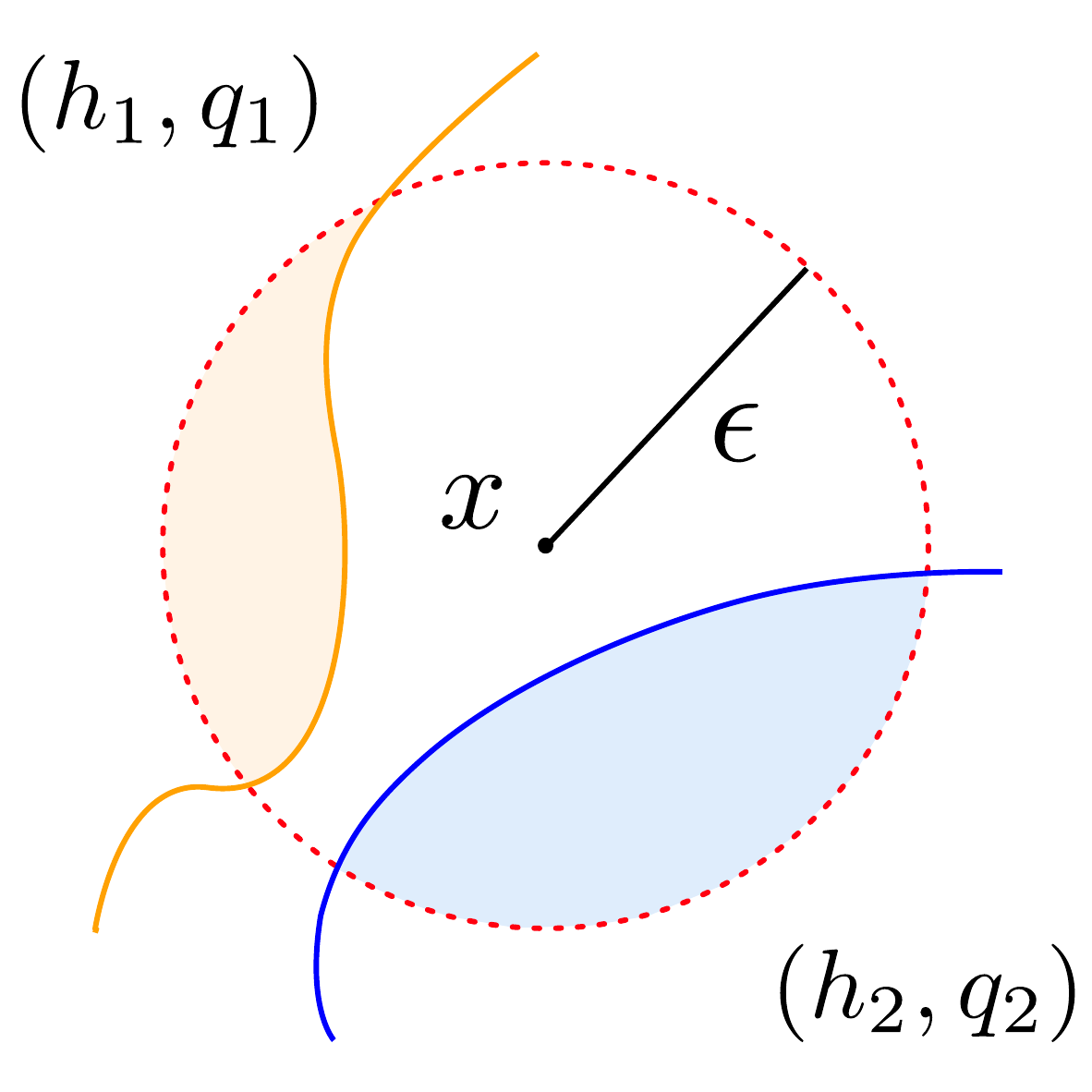} \caption{(c)}
        \label{fig:mixtures-lca-two_classifiers_regions_mp}
      }
    \endminipage\hfill
    \minipage{0.25\linewidth}%
    \centering
      {
        \includegraphics[scale=\figscale]{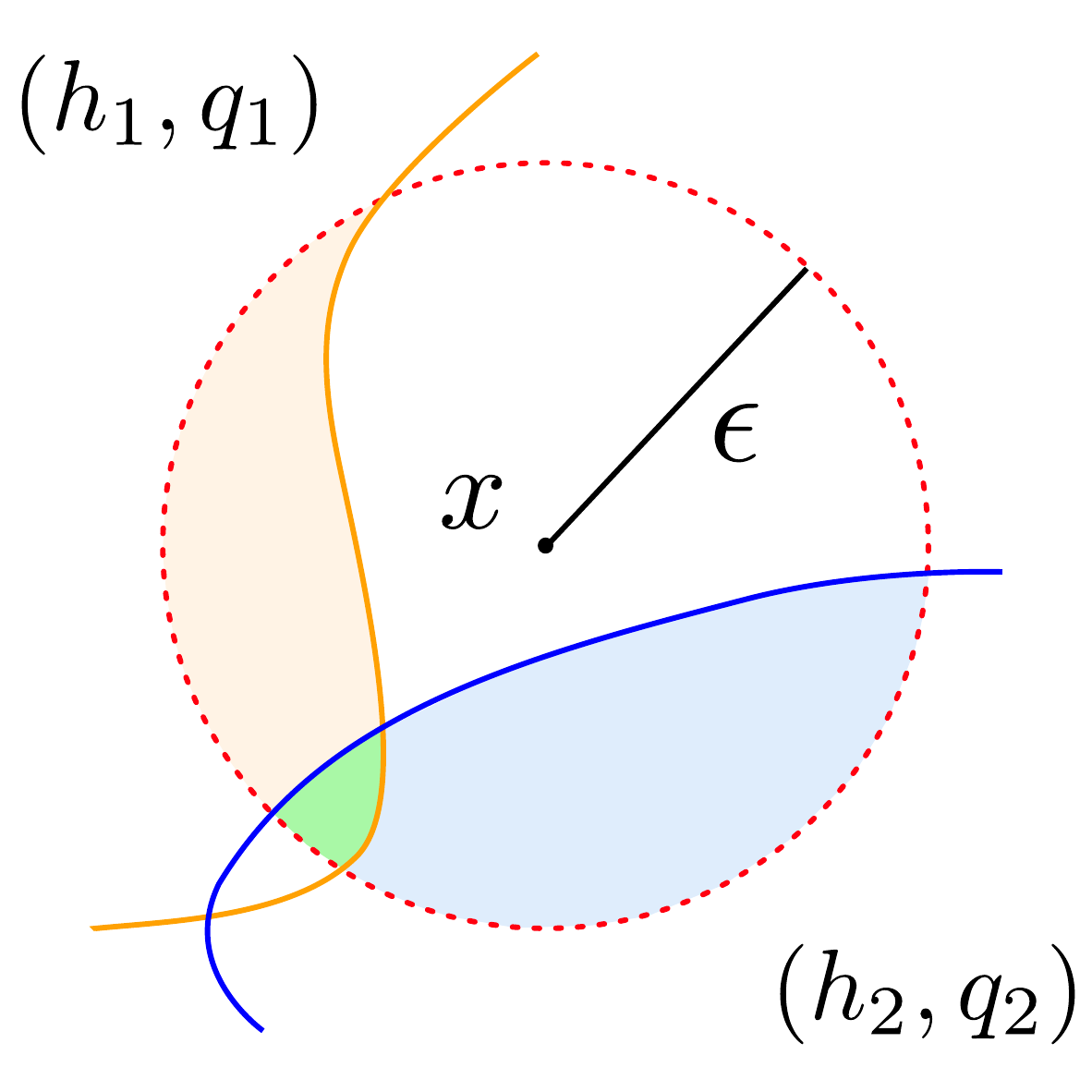} 
        \caption{(d)}
        \label{fig:mixtures-lca-two_classifiers_regions_intersection}
      }
    \endminipage
    \addtocounter{figure}{-1}
    \addtocounter{figure}{-1}
    \addtocounter{figure}{-1}
    \addtocounter{figure}{-1}
    \captionsetup{width=1.0\linewidth, labelformat=original}
    \caption{Four possible configurations for a mixture with two classifiers $h_1$ and $h_2$ assuming $x$ is correctly classified by both. a) Both $h_1$ and $h_2$ are non-vulnerable, the mixture is robust. b) Only one classifier is vulnerable. c) Both $h_1$ and $h_2$ are vulnerable, but they can not be attacked simultaneously. d) Both $h_1$ and $h_2$ are vulnerable, and they can be attacked on the same region. Best viewed in color.}
    \label{fig:mixtures-lca-two_classifiers_regions}
    \vspace{0.5cm}
\end{figure*}

\begin{figure}[ht]
  \centering
  \captionsetup{width=1.0\linewidth}
    \includegraphics[width=0.75\linewidth]{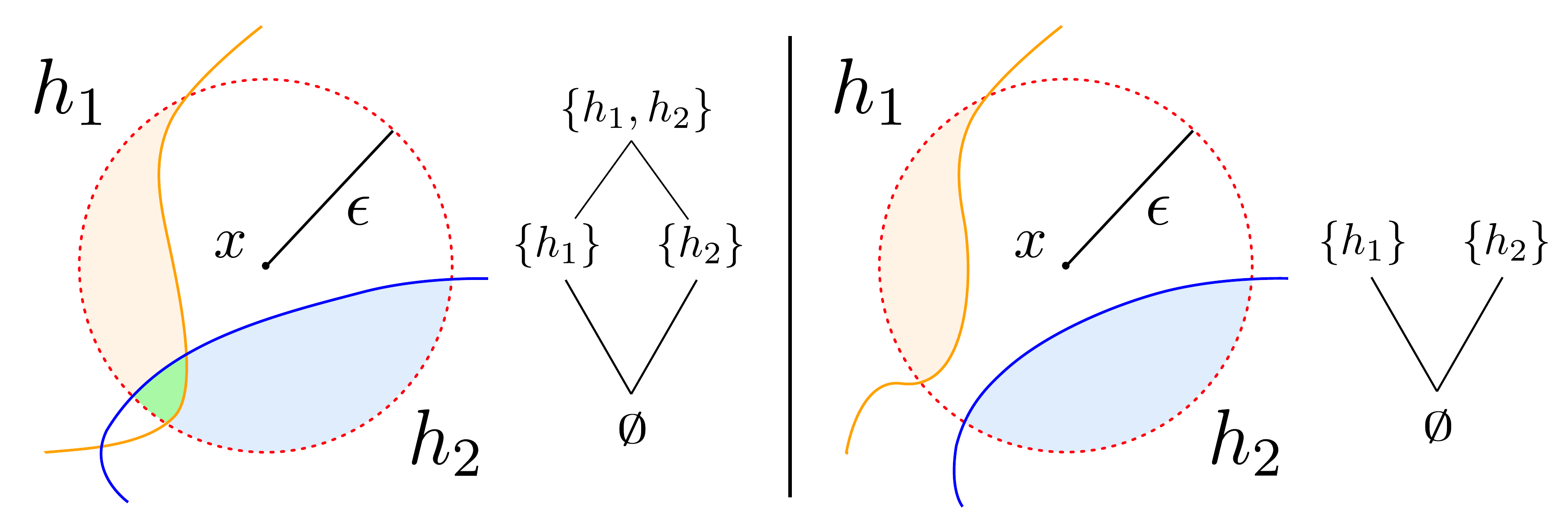}
  \caption{Examples of adversarial lattice for two of the configurations described in Figure \ref{fig:mixtures-lca-two_classifiers_regions}. Best viewed in color.}.
  \label{fig:adv_semilattice-two-configurations}
\end{figure}
\newpage

\subsection{LCA algorithm for multiclass differentiable classifiers} \label{app:lca:algo:multi}
\begin{algorithm}[h!]
\caption{LCA for multiclass classifiers}\label{alg:multiclass_attack}
\begin{algorithmic}[1]
\REQUIRE Set of $m$ classifiers $h$ in some order $(h_1 ,\dots, h_m )$. Starting point $(x, y)$. 
$T$ number of iterations and $\eta$ step size for $\mathrm{PGD}$.
\STATE Initialize pool $\mixset_{\text{pool}} = \emptyset$, $\delta = 0$
\FOR{$ i = 1, 2 \cdots, m$}\label{algo:forloop}
 \STATE $\mixset_{\text{pool}} = \mixset_{\text{pool}} \cup \{ h_i \}$ \label{algo:add_to_pool}
    
   \STATE Attack $\mathrm{SRH}(\mixset_{\text{pool}}, \cdot, y)$ starting at $x + \delta$ with $\mathrm{PGD} \left( T, \eta \right)$ to find new perturbation $\hat{\delta}$ \label{algo:attack_current} 
    \IF{$\zeroOneloss(\mixture, x+\hat{\delta}, y) > \zeroOneloss(\mixture, x+\delta, y)$}
        \STATE $\delta = \hat{\delta}$ \label{algo:update_point}
    \ENDIF
    \STATE Recompute pool $\mixset_{\text{pool}}$ according to current $\delta$       
\ENDFOR
\STATE \textbf{return} $x + \delta$\label{algo:return-end}
\end{algorithmic}
\label{algo_multiclass}
\end{algorithm}

\section{Proofs} \label{app:proofs}
\subsection{Proof of Theorem \ref{thm: hardness maxfls}}
\label{app:proof-hardness-maxfls}
\hardnessmaxfls*
\begin{proof}
    To show NP-hardness of our problem, we will use a reduction involving the NP-hard \verb|MaxFLS|~\cite{amaldi1995complexity} problem defined as follows:
    \begin{itemize}
    \item \textbf{Input of \texttt{MaxFLS}: }A set of $m$ pairs $\left\{ \left(\theta_{i},b_{i}\right)\right\} _{i=1}^{m}$
    where $\theta_{i}\in\mathbb{R}^{d}$and $b_i \in \mathbb{R}$, and a value
    $\alpha > 0$
    \item \textbf{Output of \texttt{MaxFLS}:} \verb|Yes| if there exists a vector $\delta\in\mathbb{R}^{d}$
    such that at least a fraction $\alpha$ of inequalities among the set
    $\left\{ \delta^{\top}\theta_{i}+b_{i}\ge0:i\in[m]\right\} $ are
    satisfied
    \end{itemize}
    Let us consider an instance $\left(\left\{ \left(\theta_{i},b_{i}\right)\right\} _{i=1}^{m},\alpha\right)$
    of \verb|MaxFLS|. The set $\left\{ \left(\theta_{i},b_{i}\right)\right\} _{i=1}^{m}$
    defines an arrangement of hyper-planes in $\mathbb{R}^{d}$ which can be thought as linear classifiers, and this
    arrangement partitions the space $\mathbb{R}^{d}$ into regions. Let
    $x=0$. Clearly, we can always find a value $\epsilon$ such that
    each region has a non-empty intersection with the ball $B_{p}(x, \epsilon)$.
    Let $y=0$ and $\mixweights=\left(\frac{1}{m}\ldots\frac{1}{m}\right)$, and define $\mixture$ as the mixture of the following binary classifiers
    \begin{equation*}
        h_i(x)=\mathds{1}\left\{ \theta_{i}^{\top}x+b_{i}\ge0\right\}, \text{ for } i\in[m],
    \end{equation*}
    with weights $\mixweights$. Then, $\zeroOneloss\left(\mixture,x+\delta,y\right)=\frac{1}{m}\sum_{i\in[m]}\mathds{1}\left\{ \delta^{\top}\theta_{i}+b_{i}\ge0\right\} $.
    Clearly, the value $\zeroOneloss\left(\mixture,x+\delta,y\right)$
    is equal to the fraction of satisfied inequalities among the set $\left\{ \delta^{\top}\theta_{i}+b_{i}\ge0:i\in[m]\right\} $.
    Thus, $\zeroOneloss\left(\mixture,x+\delta,y\right)\ge\alpha$
    if and only if at least an $\alpha$ proportion of inequalities in \verb|MaxFLS| are satisfied. Thus, if a polynomial time algorithm was able to solve the problem of finding an optimal adversarial attack for a mixture, we could also use it to solve the \verb|MaxFLS| problem in polynomial time, which is a contradiction, as \verb|MaxFLS| is NP-hard~\cite{amaldi1995complexity}. Thus, our problem is also NP-hard.
\end{proof}

\subsection{Proof of Lemma \ref{lemma:climb_step}}
\label{app:proof-climb-step}
\lemmaLcaMaximal*
\begin{proof}
    (See \cite[Appendix F]{perdomo2019robust})

Note that the hypothesis that there exists $x'$ adversarial to all $h \in \mixset$ means that $\mathrm{SRH}(\mixset, x', y) = 0$. This is a global minimum, as $\mathrm{SRH}$ is a non-negative function.
Also note that $\mathrm{SRH}$ as a function of $x$ can only take a finite number of values, the smallest positive one being $\frac{1}{m}$. Applying \cite[Theorem 3]{perdomo2019robust} with $\delta < \frac{1}{m}$, we conclude that running $\mathrm{PGD}$ for $T > \epsilon^2 \cdot m^2$ steps with step size $\eta = \frac{\epsilon}{\sqrt{T}}$ returns a solution $x''$ such that 
\begin{equation*}
  \mathrm{SRH}(\mixset, x'', y) - \mathrm{SRH}(\mixset, x', y) = \mathrm{SRH}(\mixset, x'', y) < \frac{1}{m}.
\end{equation*}
This implies that $\mathrm{SRH}(\mixset, x'', y) = 0$.

Note that in the actual implementation of  $\mathrm{LCA}$ (Algorithm \ref{alg:binary_linear_attack}), we will perform $\mathrm{PGD}$ multiple times with a subset of models that can be of size at most $m$. That means that we can take $T > \epsilon^2 \cdot m^2$ and $\eta < \frac{\epsilon}{\sqrt{T}}$, and these parameters would work for all the $\mathrm{PGD}$ runs during the whole execution of the algorithm. 
\end{proof}

\subsection{Proof of Theorem \ref{thm:binary_maximality}}
\label{app:proof-binary-maximality}
\thmLcaMaximality*
\begin{proof}
    Choose parameters $T$ and $\eta$ such that Lemma \ref{lemma:climb_step} holds.
    
    Denote $\mixset_i$ the pool of fooled classifiers at each step $i$ of the $m$ steps in the outer loop of Algorithm \ref{alg:binary_linear_attack}, and $x_i$ the adversarial attack at each step $i$.
    By construction, $\mixset_i \subseteq \mixset_{i+1}$, $\mixset_0 = \emptyset$ and $\mixset_m$ is the final pool of classifiers fooled by $x_m$, the output of Algorithm \ref{alg:binary_linear_attack}. 
    
    Suppose by contradiction that Algorithm \ref{alg:binary_linear_attack} returns $x_m$ that attacks a subset of classifiers $\mixset_m$ that is not maximal in the adversarial lattice of $\mixture$ at $(x, y)$. 
    
    By definition of maximality, this means there exists some classifier $h' \in \mixset \setminus \mixset_m$ such that $\mathrm{CV}(\mixset_m  \cup \{h'\}) \neq \emptyset$.
    
    Let $j$ be the step at which $h'$ was considered. At step $j-1$ of the algorithm, the pool $\mixset_{j-1}$ consists of classifiers that are all vulnerable. Given that $\mixset_{j-1} \cup \{h'\} \subset \mixset_m \cup \{h'\}$, we have that $\mathrm{CV}(\mixset_{j-1} \cup \{h' \}) \supset \mathrm{CV}(\mixset_m \cup \{h'\}) \neq \emptyset$.
    
    As $\mathrm{CV}(\mixset_{j-1} \cup \{h'\}) \neq \emptyset$, at step $j$, Lemma \ref{lemma:climb_step} ensures that the $\mathrm{PGD}$ step for attacking the pool of classifiers $\mixset_{j-1} \cup \{h'\}$ returns $x_j$ such that
    \begin{equation*}
        \mathrm{SRH}(\mixset_{j-1} \cup \{h'\}, x_j, y) = 0,
    \end{equation*}
    meaning that $h'$ would be added to the pool, \textit{i.e} $\mixset_j = \mixset_{j-1} \cup \{h'\}$. This implies that $h' \in \mixset_m$, which is a contradiction.
    
    This contradiction arises from supposing that $\mathrm{CV}(\mixset_m)$ was not maximal, which concludes the proof.
\end{proof}

\section{Synthetic experiments} 
\label{app:toy}

In this section we provide more details on the synthetic experiments presented in Section \ref{sec:experiments:synthetic}. The first experiment is the one in $\mathbb{R}^2$ with two classifiers that produced Figure \ref{fig:attack:apgd-arc-against-optimal} in which we vary the angle between the normal vectors of the two linear classifiers and leave their distance to the center point fixed. The second experiment is the one in which we sample $m$ linear models in $\mathbb{R}^d$ with different values for the dimension $d$, and the linear classifiers are sampled with different distances to the center point.

\subsection{ARC and EOL-PGD against two classifiers in the plane at the same distance from the origin}

These are the parameters used for the experiment that produced \ref{fig:attack:apgd-arc-against-optimal}. The LCA parameters are
\begin{itemize}
    \item Norm: $\ell_2$.
    \item Attack budget $\epsilon$: 1.
    \item Number of steps $T$ for EOL-PDG: 100.
\end{itemize}
ARC in the binary linear setting has no parameters to tune.

\subsection{LCA, EOL-PGD and ARC with different number of classifiers with respect to the distribution of the bias of the linear classifiers}

\begin{figure}[ht]
    \centering
      \includegraphics[scale=0.4]{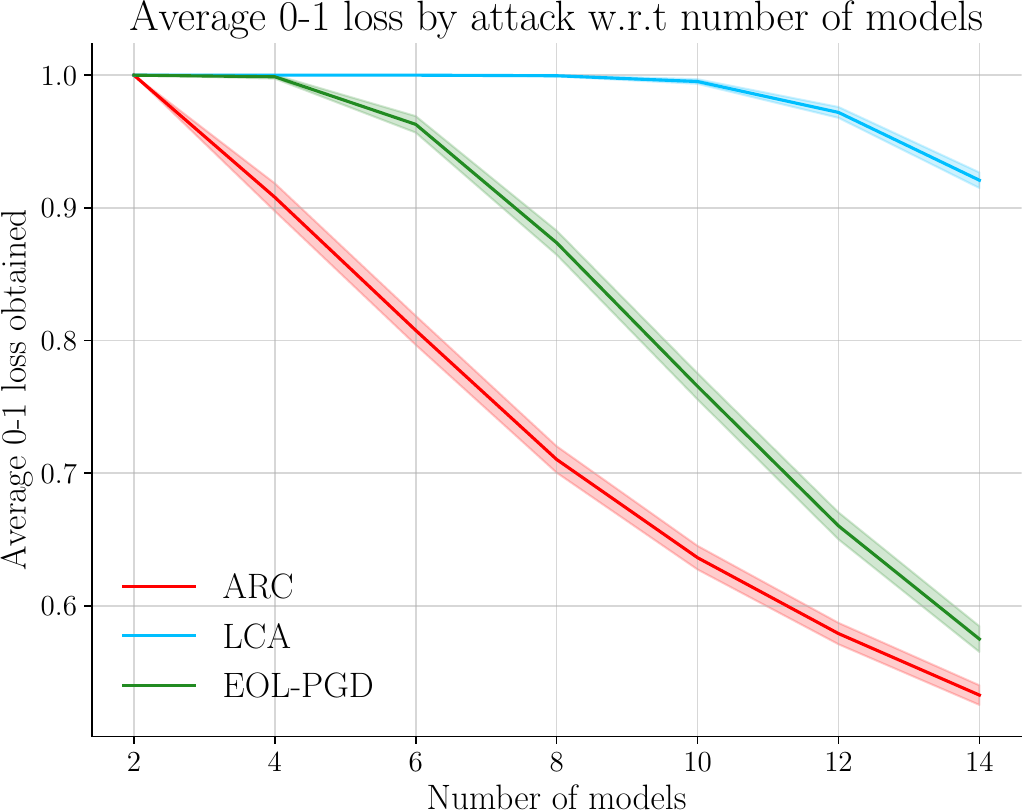}
    \captionsetup{width=1.0\linewidth, skip=8pt}
    \caption{Biases sampled from $|\cal N(0.25, 0.05)|$. Classifiers are very close to the center point (low $\alpha$), and they are almost all at the same distance (low $\beta$). In this particular case LCA performs very well, being almost always optimal. EOL-PGD also performs really well up to 6 models, which means that until that point, almost all linear classifiers are well aligned, and the common vulnerability region can be found by attacking the average loss. ARC on the other hand degrades sharply as the number of models increases.}
    \label{fig: 2d score num models plot d128 a025 b005}
    \vspace{1cm}
\end{figure}

\begin{figure}[!ht]
    \centering
      \includegraphics[scale=0.4]{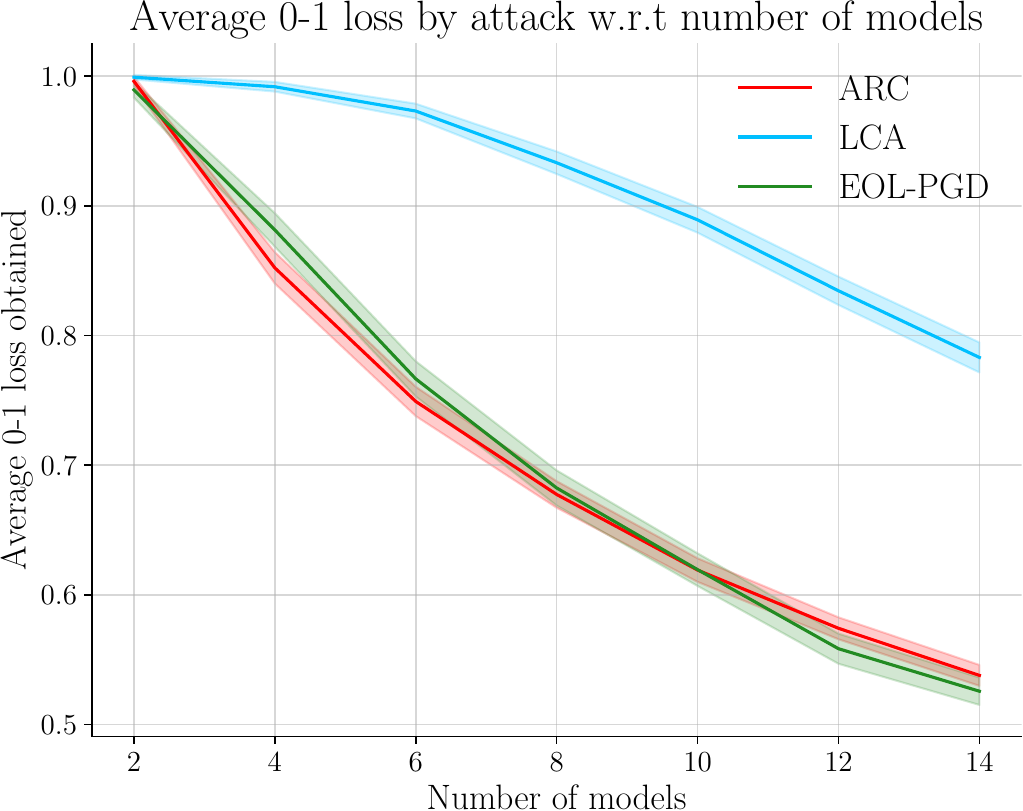}
    \captionsetup{width=1.0\linewidth, skip=8pt}
    \caption{Biases sampled from $|\cal N(0.25, 0.2)|$. Classifiers are on average very close to the point but with a higher variance in their distance.}
    \label{fig: 2d score num models plot d128 a025 b02}
    \vspace{1cm}
\end{figure}

We tested EOL-PGD, ARC and LCA against randomly sampled mixtures of different sizes. To control the difficulty of the setting, we introduced the parameters $\alpha$ and $\beta$, which control the average and the variance of the distance from the center point $x$ to the decision boundaries of the linear classifiers, respectively. Figures \ref{fig: 2d score num models plot d128 a025 b005}, \ref{fig: 2d score num models plot d128 a025 b02} and \ref{fig: 2d score num models plot d128 a075 b02} are similar to Figure \ref{fig: 2d score num models plot} in the main text but with other values of $\alpha$ and $\beta$, where it can be seen that the hardness of crafting an attack can change drastically depending on the configuration of the classifiers around the point. 
Recall that $\alpha$ is the average distance from the center point to the decision boundaries, and $\beta$ is the variance of the distances.
In all cases, note that the $y-$axis values are not the same, and have been adapted for visual purposes.

\begin{figure}[!ht]
    \centering
      \includegraphics[scale=0.4]{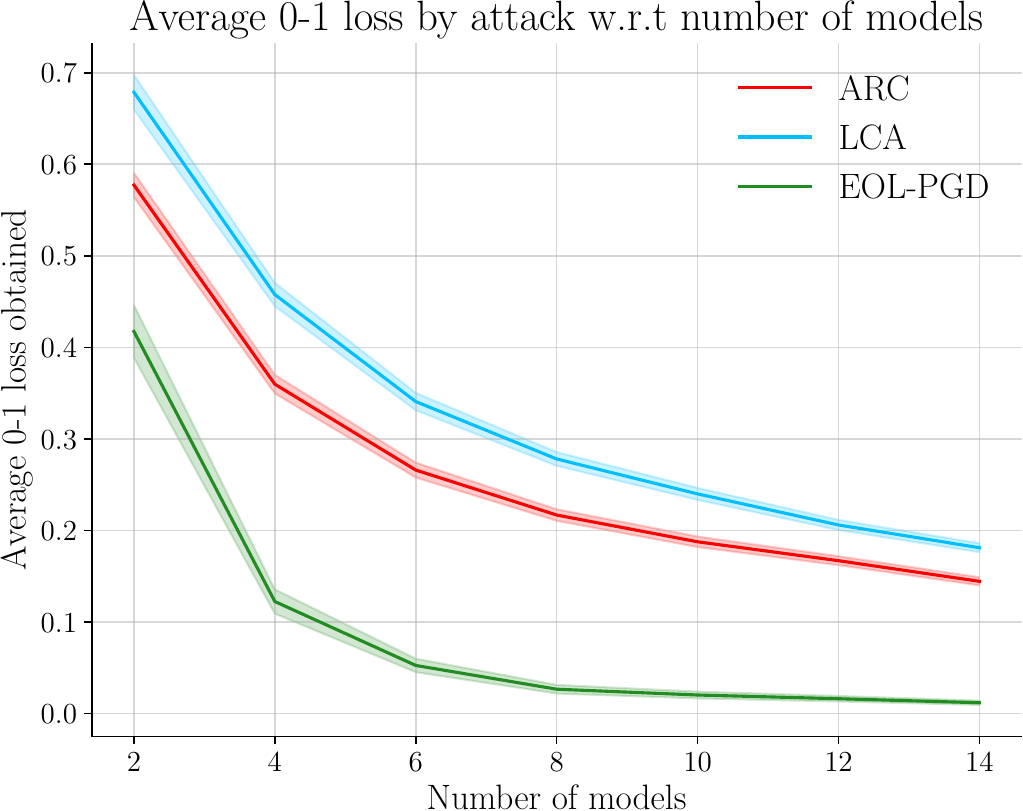}

    \captionsetup{width=1.0\linewidth, skip=8pt}
    \caption{Biases sampled from $|\cal N(0.75, 0.2)|$. This configuration is hard because classifiers are close to the boundary of the $\epsilon-$ball (far from $x$), and the probability of intersection is less than in the cases seen before. LCA is on average superior, but the gap between LCA, ARC and EOL-PGD is smaller.}
    \label{fig: 2d score num models plot d128 a075 b02}
    \vspace{1cm}
\end{figure}

In order to check that LCA was performing better than ARC and EOL-PGD, we computed the average over 100 runs of the ratio between the score obtained by each attack and the score that LCA obtained in the same configuration. Figure \ref{fig: 2d ratio num models plot d128 a075 b02} shows the evolution of such ratio with respect to the number of models. The colored area around the lines corresponds to a 99\% confidence interval for the average ratio. We can see that LCA is superior by a rate that tends to stabilize as the number of models increase.

\def\figscale{0.3}
\begin{figure}[!ht]
    \centering
    \captionsetup{width=.31\linewidth,labelformat=empty,skip=8pt,justification=centering}
    \minipage{0.31\linewidth}%
    \centering
      {
        \includegraphics[width=\textwidth]{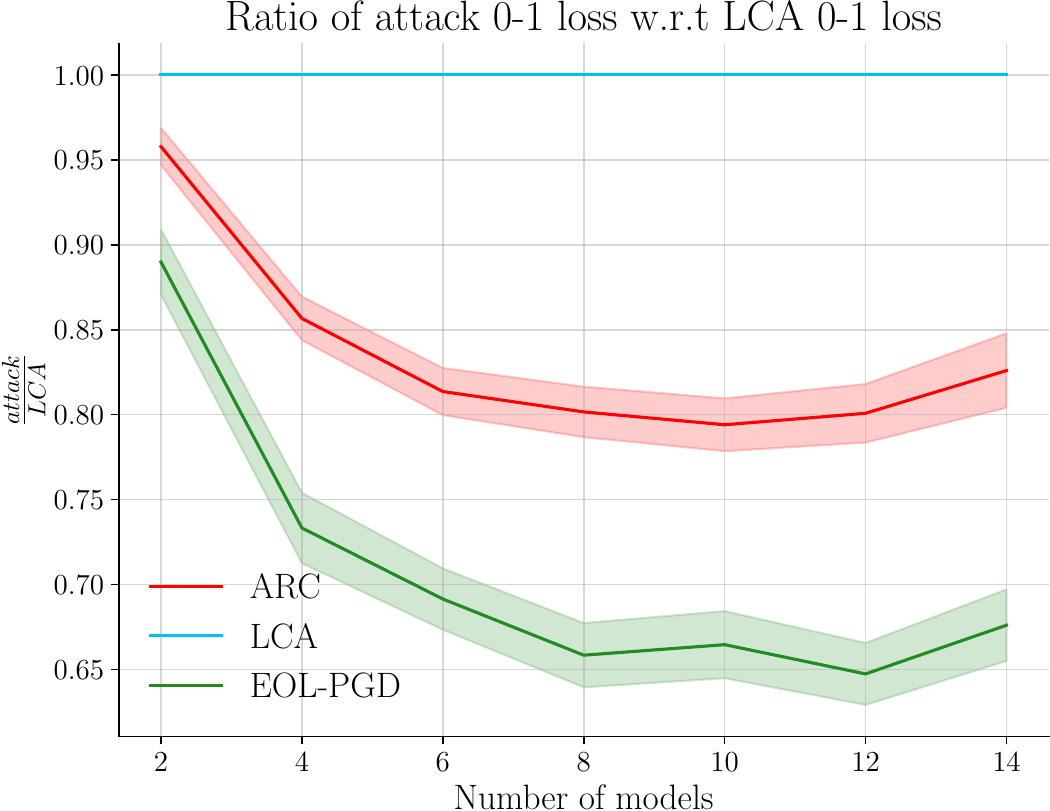}
        \caption{$|\cal N(0.25, 0.5)|$}
        \label{fig: ratio a025 b05}
      }
    \endminipage\hfill
    \minipage{0.31\linewidth}
    \centering
      {
        \includegraphics[width=\textwidth]{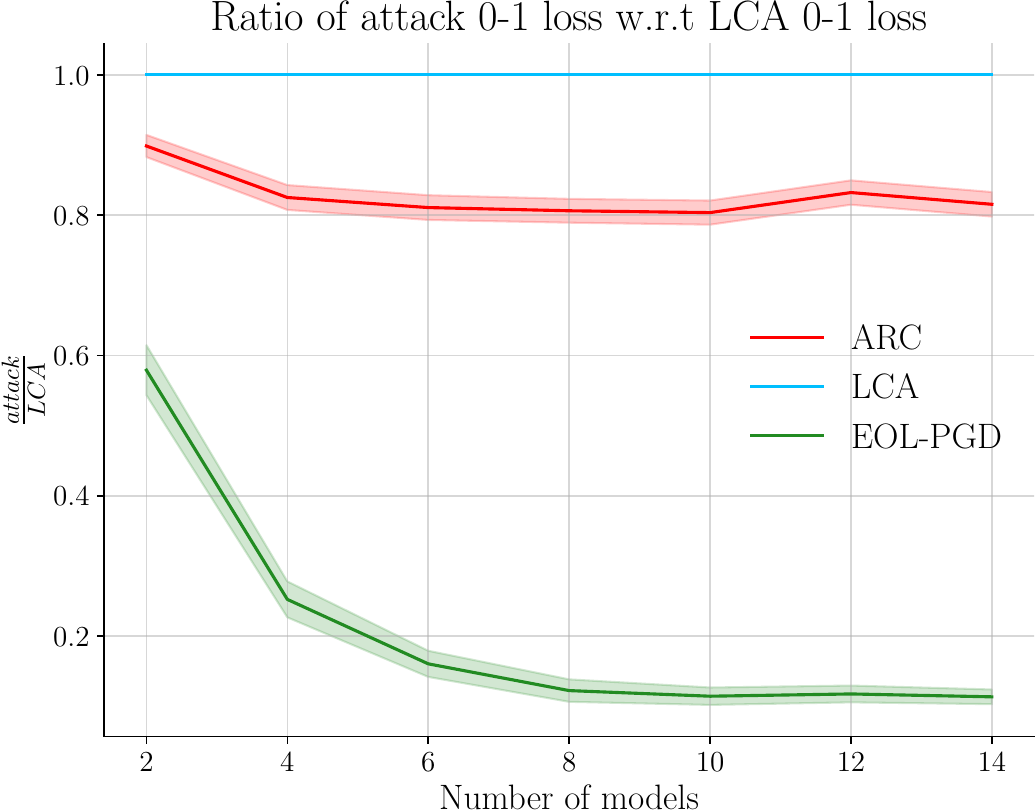}
        \caption{$|\cal N(0.75, 0.2)|$}
        \label{fig: ratio a075 b02}
      }
    \endminipage\hfill
    \minipage{0.31\linewidth}
    \centering
      {
        \includegraphics[width=\textwidth]{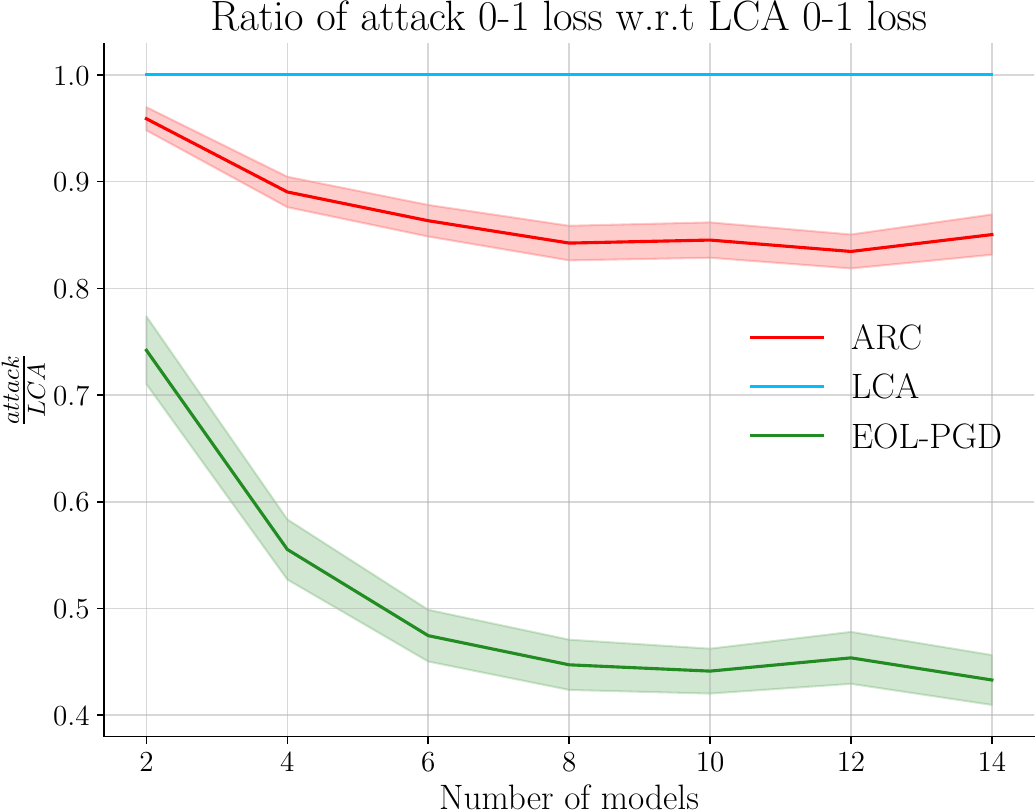}
        \caption{$|\cal N(0.75, 0.5)|$}
        \label{fig: ratio a075 b05}
      }
    \endminipage
    \addtocounter{figure}{-1}
    \addtocounter{figure}{-1}
    \addtocounter{figure}{-1}

    \captionsetup{width=0.95\linewidth,labelformat=original}
    \caption{Ratios for different values of $\alpha$ and $\beta$. $y-$ axis are adapted for visual purposes. }
    \label{fig: 2d ratio num models plot d128 a075 b02}
    \vspace{1cm}
\end{figure}

Here are the parameters used for all these experiments:

\textbf{General parameters:}
\begin{itemize}
    \item Norm (threat model): $\ell_2$.
    \item Attack budget $\epsilon$: 1.
    \item ARC for linear classifiers has no extra parameters. That is why it is extremely fast compared to EOL-PGD and LCA.
    \item \verb|numpy| seed was set to 42
    \item {Weights $\mixweights^{(i)}$ of each random mixture were also sampled randomly as follows:
        \begin{itemize}
            \item Sample $m$ i.i.d standard normal values $z_i$.
            \item Compute the softmax with temperature $t=10$, \textit{i.e} $\mixweights^{(i)} = \frac{\exp{(z_i/10)}}{\sum_j \exp{(z_j/10)}}$.
        \end{itemize}
    }
\end{itemize}

\textbf{Parameters for LCA:}
\begin{itemize}
    \item Number of steps $T$ for inner PDG: $\max(2\cdot m^2, 200)$.
    \item Step size $\eta$ for inner PDG: $\frac{\epsilon}{\sqrt{T}}$.
\end{itemize}

\textbf{Parameters for EOL-PGD:}
\begin{itemize}
    \item Number of steps $T$ for inner PDG: $\max(2\cdot m^2, 200)$.
    \item Step size $\eta$ for inner PDG: $\frac{\epsilon}{\sqrt{T}}$.
\end{itemize}

\section{CIFAR-10 experiments} \label{app:cifar}

\subsection{Main results}

For the experiments with the pre-trained models available in the DVERGE repository\footnote{https://github.com/zjysteven/DVERGE/tree/main}, we attacked the models with different versions of EOL-PGD, ARC and LCA, trying to cover a reasonable range of hyperparameters for all attacks such as the step size $\eta$ and the number of iterations $T$. For the number of iterations, we tested 50, 100 and 500 for EOL-PGD, and 50, 100 and 200 for ARC. LCA was only tested with 50, to avoid making an unfair comparison. We tested LCA with a constant step size, and also with a changing step size that is divided by 2 at the iteration 40. The best results were used for the results in Table \ref{table: cifar-10 results}.

Here is a summary of the parameters used for the main results in Table \ref{table: cifar-10 results}:

\textbf{General parameters:}
\begin{itemize}
    \item Norm: $\ell_{\infty}$.
    \item Attack budget $\epsilon$: 0.01 or 0.03.
\end{itemize}

\textbf{Parameters for LCA:}
\begin{itemize}
    \item Number of steps $T$ for inner PDG: 50.
    \item {Step size $\eta$ for inner PDG:
        \begin{itemize}
            \item For $\epsilon = 0.01$: 0.0005, 0.001, 0.002
            \item For $\epsilon = 0.03$: 0.0015, 0.003, 0.006
        \end{itemize}
    }
\end{itemize}

\textbf{Parameters for EOL-PGD:}
\begin{itemize}
    \item Number of steps $T$: 50, 100 and 500.
    \item {Step size $\eta$:
        \begin{itemize}
            \item For $\epsilon = 0.01$: 0.0015, 0.002, 0.006
            \item For $\epsilon = 0.03$: 0.0015, 0.006
        \end{itemize}
    }
\end{itemize}

\textbf{Parameters for ARC:}
\begin{itemize}
    \item Number of steps $T$: 50, 100, 200.
    \item {Step size $\eta$:
        \begin{itemize}
            \item For $\epsilon = 0.01$: 0.0005, 0.001, 0.002, 0.006
            \item For $\epsilon = 0.03$: 0.0015, 0.003, 0.006
        \end{itemize}
    }
\end{itemize}

\subsection{Comparing strong versions of the attacks}
We wanted to make sure that the gap between EOL-PGD, ARC and LCA was not influenced by the simple setting we considered for the main experiments in Table \ref{table: cifar-10 results}. We also wanted to test stronger versions of LCA to evaluate the possibility of further improvement. 

Firstly, we launched stronger versions of both ARC and LCA against the model DV+AT/5 in the $\epsilon = 0.03$ setting. We added random initialization and 5 restarts to both attacks, and tested with $T \in \{50, 100, 250\}$. For LCA, we also tested dividing the step size of the inner PGD by 2 at iterations $\frac{T}{2}$ and $\frac{3T}{4}$.  

From all the different evaluations, we report the lowest robust accuracy (strongest attack) and compare it to the results obtained in Table \ref{table: cifar-10 results}.

\begin{table}[!ht]
\setlength{\aboverulesep}{0pt}
\setlength{\belowrulesep}{0pt}
    \caption{Mixture robustness evaluation of DV+AT/5 against stronger versions of ARC and LCA.}
      \centering
            \begin{tabular}{|x{1.5cm}|x{1.5cm}|x{1.5cm}|}
            \toprule
            Attack & From Table \ref{table: cifar-10 results} & Stronger \\
            \midrule
            ARC & 44.6\% & 43.6\%  \\
            LCA & 39.9\% & 36.5\%  \\
            \bottomrule
           \end{tabular}
           \label{table: cifar-10 arc_lca strong}
\end{table}

Including random initialization and restarts only changed the results for ARC by $1\%$ with respect to the results shown in Table \ref{table: cifar-10 results}. On the other hand, LCA improved by $3.4\%$, showing that the robustness evaluation of LCA could further improve with more hyperparameter tuning.

Secondly, we performed a more particular experiment varying only the number of iterations, similar to a convergence check when verifying that a proposed defense is not overestimating the reported robustness due to the non-convergence of the attack \cite{tramer2020adaptive}. Again, we used the model DV+AT/5 in the $\epsilon = 0.03$ setting. We set the step size to $0.006$ for all attacks, no random initialization, and LCA was tested with and without the reducing step size at iterations $\frac{T}{2}$ and $\frac{3T}{4}$ by half.  

It can be seen that further increasing the number of iterations up to 1000 does not reduce the robustness evaluation of EOL-PGD or ARC by more than 0.2\%. On the other hand, for this fixed setting, LCA with the reducing step size obtains an improvement of 0.8\%, suggesting again that the reasonable version of LCA that we used for the main results, that already beats EOL-PGD and ARC, can still be enhanced.

\begin{table}[!ht]
\setlength{\aboverulesep}{0pt}
\setlength{\belowrulesep}{0pt}
    \caption{Convergence check. Robustness evaluation with an increasing number of iterations. LCA$^\dag$ corresponds to the version of LCA with the reducing step size.}
      \centering
            \begin{tabular}{|x{1.5cm}||x{1.5cm}|x{1.5cm}|x{1.5cm}|}
            \toprule
             \multirow{2}{*}{Attack} &  \multicolumn{3}{c|}{$T$}\\
            \cmidrule{2-4}
              & 100 & 500 & 1000 \\
            \midrule
            EOL   &  42.0\% & 41.9\% & 41.9\% \\
            ARC    &  44.1\% & 44.0\% & 43.9\% \\
            LCA    &  38.5\% & 38.0\% & 38.0\% \\
            LCA$^\dag$    &  38.5\% & 37.9\% & 37.7\% \\
            
            \bottomrule
           \end{tabular}
           \label{table: cifar-10 convergence}
\end{table}


\paragraph{Optimality gap: the impact of the number of restarts.}
Given that $\mathrm{PGD}$ has no optimality guarantee in the case of general multiclass differentiable classifiers, we cannot compute the real optimality gap of  LCA. We can, however, compare  LCA against a much stronger version of it that tries all the possible orderings of classifiers to perform the lattice climb. Recall that the ordering in which the classifiers are considered determines the path that is used to climb the lattice, and thus the maximal region that is attained. In practice, this ordering is even more important for Algorithm \ref{alg:multiclass_attack}, in which the pool of classifiers is adapted depending on the score, and can therefore change in a non-increasing manner.

To have a better understanding of the limits of  LCA, we launched  LCA with different numbers of random restarts, ranging from 1 (normal LCA) to 10, which means that 10 different random orderings are tested, and at the end of each ordering, only the best perturbation is kept in a greedy way. As an optimality baseline, we launched a brute-force version of LCA, denoted  LCA$_{\mathrm{all}}$, that tests all the possible orderings and keeps, at each time, the best perturbation. We used the same mixtures from Table \ref{table: cifar-10 results}, but only for the models AT/5, DV+AT/5, BARRE/5 and MR/5 that showed a good level of robustness. As all these mixtures have 5 sub-models, there is a total of 120 possible orderings tested by  LCA$_{\mathrm{all}}$.

\def\figscale{0.45}
\begin{figure}[ht]
    \centering
    

    \includegraphics[scale=\figscale]{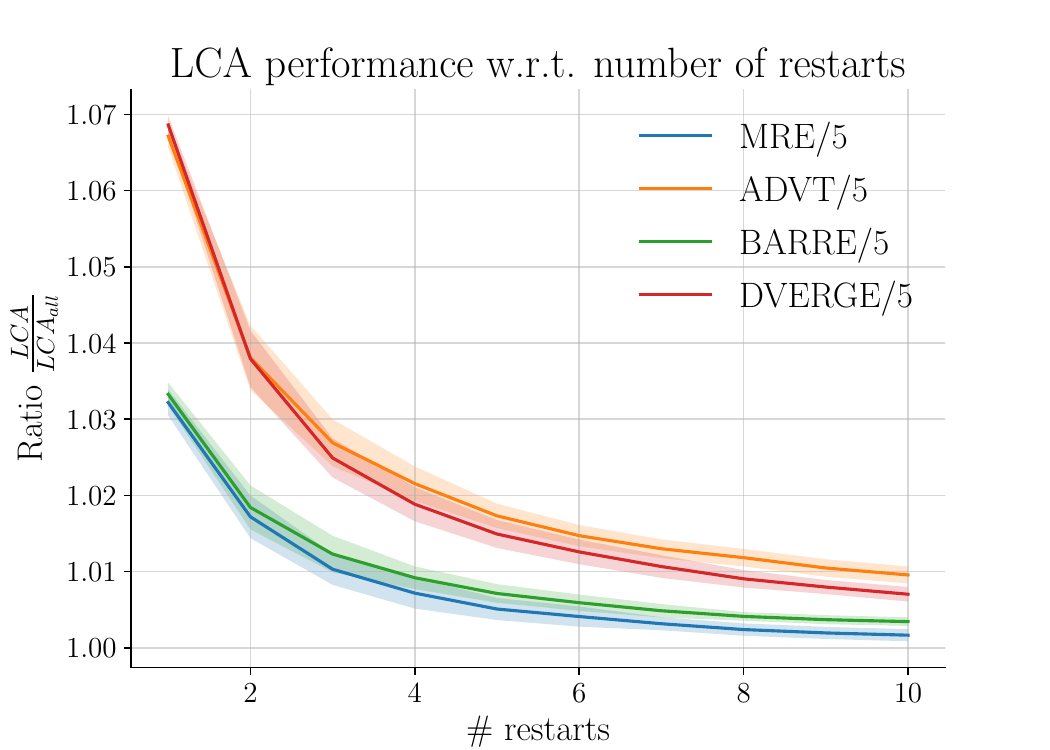} 

    \caption{Impact of restarts in the performance of  LCA. The curves show the ratio of the accuracy after attack by  LCA and the brute force  LCA$_{\mathrm{all}}$ (lower the better), averaged over 20 runs with different seeds. Colored regions represent three standard deviations from the average. Best viewed in color.}
    \label{fig: optimality gap}
\end{figure}
Figure \ref{fig: optimality gap} shows the ratio of the robust accuracy obtained by LCA and  LCA$_{\mathrm{all}}$ as a function of the number of restarts. The reported values are the average over 20 different seeds, and the shaded region around each curve represents three standard deviations from the mean. It can be seen that with 10 restarts, the ratio between  LCA and  LCA$_{\mathrm{all}}$ is of at most 1\% which translates to an absolute accuracy gap of at most 0.38\% for the model AT/5 that has the highest gap. This experiment also shows that if more computation time is available to afford random restarts, LCA can improve even further with respect to the main results shown in Table \ref{table: cifar-10 results} where it was already superior to ARC and EOL-PGD. In our experiment, with 5 restarts for models AT/5 and DV+AT/5,  LCA can decrease the accuracy by an extra 2\% with respect to the baseline without restarts. For the models MR/5 and BARRE/5, the improvement is of 1\%. Note that all the other parameters for  LCA were left unchanged from the settings used for the main results (50 iterations for the inner $\mathrm{PGD}$, no random initialization of the perturbation).

\section{On adversarial training for mixtures of classifiers} \label{app:adversarial-training-mixtures}

A natural second step, after proposing an attack, is to see if it is beneficial for adversarial training. Our standard training for mixtures optimizes the sum of losses, which makes the training independent even if models are trained at the same time (not sequentially like in \cite{dbouk2023robustness}). To add dependence, we test a vanilla adversarial training with a mixture attack: the attack is crafted for the mixture, which breaks the independence in the training process.

We tested adversarial training with EOL-PGD, ARC, LCA and vanilla PGD on the ensemble that averages logits, in CIFAR-10, with different architectures for our sub-models. We tested with very small convolutional networks (CNN), with the ResNet20 models used in the experiments in Table \ref{table: cifar-10 results}, and also with a bigger ResNet110. All these experiments were carried out with a different threat model ($\ell_2$ with $\epsilon = 0.5$), so they are not comparable to the results in Table \ref{table: cifar-10 results}.

\begin{table}[!ht]
\setlength{\aboverulesep}{0pt}
\setlength{\belowrulesep}{0pt}
    \caption{Clean and robust accuracy of mixtures with different base models. Mixtures have been trained with adversarial training using different attacks during training under the $\ell_{2}$ threat model with $\epsilon = 0.5$ as in standard practice. Higher is better (more robust). Robust accuracy reported is the minimum expected accuracy under any of the tested attacks (EOL-PGD, ARC, LCA, LOE-AutoAttack and LOE-PGD).}
      \centering
            \begin{tabular}{|x{1.9cm}|x{2cm}|c|c|}
            \toprule
            Model & Attack for AT &  Natural & Robust \\
            \midrule
            \multirow{3}{*}{CNN} & EOL-PGD & 59.9\% & 34.1\% \\
                                      & LOE-PGD  &  61.8\% & 32.8\% \\
                                      & LCA  &  64.0\% & 27.2\% \\
            \midrule                
            \multirow{4}{*}{ResNet20} & EOL-PGD & 81.2\% & 50.0\% \\
                                      & LOE-PGD  &  81.5\% & 43.4\% \\
                                      & LCA  &  83.1\% & 49.6\% \\
                                      & ARC  & 84.5\% & 46.5\% \\
            \midrule                
            \multirow{3}{*}{ResNet110} & EOL-PGD & 77.8\% & 47.2\% \\
                                      & LOE-PGD  &  78.2\% & 40.6\% \\
                                      & LCA  &  80.7\% & 46.3\% \\
            \bottomrule
           \end{tabular}
           \label{table: cifar-10 at mixtures}
\end{table}

We can see that, as has already been noted in the seminal paper \cite{madry2017towards}, models with very low capacity are less able to learn when faced with more complex and stronger attacks. In the case of the CNN, training with EOL-PGD produces models with an individual robustness similar to what independent adversarial training would produce. On the other hand, training with LCA produces models with drastically less robustness. This was noted for all the tested versions of LCA (from 5 to 30 iterations, and various step sizes) and even with different mixtures sizes.

When the capacity is increased, the performance of AT with EOL-PGD and LCA becomes almost on par. It is intriguing that AT with ARC and LCA yields models with higher standard accuracy. We expected the opposite, considering that LCA and ARC are supposed to be stronger attacks against mixtures. This behavior led the authors in \cite{dbouk2023robustness} to use EOL-PGD for their training method, BARRE, instead of ARC.

The problem of training a mixture that increases the robustness with respect to the best individual sub-model remains open and is an interesting track for future work.

\end{document}